\documentclass[11pt]{article}

\usepackage[final]{acl}

\usepackage{times}
\usepackage{latexsym}
\usepackage{booktabs}
\usepackage{tabularray}
\usepackage{multirow}
\usepackage{tcolorbox}
\tcbuselibrary{skins,breakable}
\usepackage{needspace}
\usepackage{listings}
\usepackage[T1]{fontenc}

\usepackage[utf8]{inputenc}

\usepackage{microtype}

\usepackage{inconsolata}

\usepackage{graphicx}

\title{IndicDBBench - Benchmarking Multilingual Text-to-SQL Capabilities in Indian Languages}

\author{
 \textbf{Aviral Dawar\textsuperscript{1}},
 \textbf{Roshan Karanth\textsuperscript{2}},
 \textbf{Vikram Goyal\textsuperscript{3}},
 \textbf{Rakhee Kulshrestha\textsuperscript{2}},
 \\
 \textbf{Sumanta Pasari\textsuperscript{2}},
 \textbf{Dhruv Kumar\textsuperscript{2}}
\\
\\
 \textsuperscript{1}Independent Researcher,
 \textsuperscript{2}BITS Pilani, Pilani, India,
 \textsuperscript{3}IIIT Delhi, New Delhi, India
\\
}

\begin{document}
\maketitle
\begin{abstract}
While Large Language Models (LLMs) have significantly advanced Text-to-SQL performance, existing benchmarks largely focus on Western contexts leaving a gap in real-world, non-Western applications. We introduce IndicDBBench, a comprehensive multilingual Text-to-SQL benchmark designed to evaluate cross-lingual semantic parsing across diverse Indic language families. The underlying schemas are derived from real-world open-data platforms (\href{https://ndap.niti.gov.in/}{NDAP} and \href{https://indiadataportal.com/}{IDP}), ensuring realistic structural complexity. IndicDBBench comprises \textbf{20 databases spanning 237 tables}. To construct complex relational structures from denormalized government data, we employ a three-agent pipeline (Architect, Auditor, Refiner), achieving high relational density (11.85 tables per database) and join depths of up to six. We generate 15,612 tasks spanning seven languages: English and six Indic languages, ensuring that queries are grounded in real database values and require multi-table reasoning. We evaluate state-of-the-art models, including DeepSeek V3.2, MiniMax M2.7, Llama 3.3, and Qwen3, across seven linguistic variants. Our results reveal a 9.0-point drop in execution accuracy from English to Indic languages, exposing an “Indic Gap” that, for the two models tested, persists even when gold tables and literal values are supplied, since this grounding raises accuracy in every language by similar amounts. The dataset and code are publicly available \href{https://anonymous.4open.science/r/multilingualText2Sql-Indic--DDCC/}{here}.
\end{abstract}

\section{Introduction}

Text-to-SQL parsing translates natural language questions into executable SQL queries, enabling non-expert users to interrogate relational databases without mastering query syntax. Driven by advances in Large Language Models (LLMs), top-model execution accuracy on Spider~\citep{spider} rose from 53.5\% to 91.2\%, while BIRD~\citep{bird} and Spider\,2.0~\citep{spider2} show that large, noisy, enterprise-grade settings remain far from solved.

A critical blind spot in this progress is its overwhelmingly \textbf{English-centric} nature: Spider, BIRD, Spider\,2.0, and WikiSQL use English-only tasks with Western-context schemas, and even multilingual extensions such as MultiSpider~\citep{multispider} fail to capture the administrative and linguistic complexities of the \textbf{Indian Sub-continent}. IndicDBBench addresses this limitation with a specialized suite that rigorously tests cross-lingual semantic parsing and structural reasoning across the diverse scripts and relational frameworks of the Indian subcontinent.

India's public data ecosystem, hosted on platforms such as \textbf{NDAP}, \textbf{IDP}, \textbf{ICRISAT}, and \textbf{IHDS}, serves as a challenging evaluation testbed. These datasets feature deep administrative hierarchies (Country $\rightarrow$ State $\rightarrow$ District $\rightarrow$ Sub-District $\rightarrow$ Block $\rightarrow$ Village), resulting in foreign-key chains with a depth of six. IndicDBBench covers Indian administrative domains that current benchmarks lack, such as household surveys and census demography. Representative examples include an 18-table health surveillance database covering routine immunization and family planning, alongside agricultural datasets using seasonal columns such as \texttt{KHARIF\_SORGHUM\_YIELD\_KG\_PER\_HA}. High-cardinality entity spaces (tables with hundreds of thousands of unique values) impose extreme schema-linking demands, particularly for Indic language queries that lack lexical overlap with English-encoded column names.
\subsection{IndicDBBench: Benchmark and Contributions}

We present \textbf{IndicDBBench}, a large-scale multilingual Text-to-SQL benchmark grounded in real Indian
administrative databases, evaluated across \textbf{seven linguistic variants}: English, Hinglish, Hindi,
Bengali, Tamil, Telugu, and Marathi. Our contributions are:

\begin{itemize}
    \item \textbf{Systematic Construction and Synthesis.} We curate \textbf{20 PostgreSQL databases (237 tables, 7.69M rows)} via a novel \textbf{three-agent judge pattern} (Architect, Auditor, Refiner), producing star/snowflake schemas with join-depths up to six that support \textbf{15,612 tasks} synthesized by a value-aware, join-enforced pipeline across seven languages, screened with COMET-QE and reviewed by native speakers on flagged and randomly sampled translations.

\item \textbf{Comprehensive Multi-Model Benchmarking.} We evaluate \textbf{Llama 3.3 70B}, \textbf{Qwen3 8B}, \textbf{MiniMax M2.7}, and \textbf{DeepSeek V3.2} across zero-shot and \textbf{DIN-SQL} prompting \citep{llama3, qwen3, minimax2026, deepseek2025, dinsql}, and test the impact of SEED evidence augmentation~\citep{seed} on cross-lingual grounding in high-cardinality environments.

\item \textbf{Characterization of the ``Indic Gap.''} We uncover a consistent \textbf{$\sim$9.0-point} global execution-accuracy drop from English to Indic variants with the most substantial deficit observed in Telugu, which exhibits a maximum decline of \textbf{$\sim$12.8 points}, as detailed in \textbf{Table~\ref{tab:performance-delta}}. Through fine-grained \textbf{error analysis}, we categorize failure modes across schema complexities and linguistic nuances, providing actionable insights for improving multilingual Text-to-SQL reasoning.
\end{itemize}

\begin{table}[t]
  \begin{center}
    \begin{small}
      \begin{sc}
        \begin{tabular}{lcc}
          \toprule
          \textbf{Language} & \textbf{Avg. EX} & \textbf{Drop (pp)} \\
          \midrule
          English  & 64.6\% & -- \\
          Hinglish & 58.3\% & 6.3 \\
          Bengali  & 56.9\% & 7.7 \\
          Hindi    & 55.5\% & 9.1 \\
          Marathi  & 54.7\% & 9.9 \\
          Tamil    & 56.3\% & 8.3 \\
          Telugu   & 51.8\% & \textbf{12.8} \\
          \bottomrule
        \end{tabular}
      \end{sc}
    \end{small}
  \end{center}
    \caption{Cross-lingual EX on IndicDBBench. Telugu exhibits the most significant accuracy reduction relative to English.}
      \label{tab:performance-delta}
  \vskip -0.1in
\end{table}

\section{Related Work}

\textbf{Text-to-SQL Benchmarks and Multilingual Transfer.} Evaluation has progressed from Spider's cross-domain challenge \citep{spider} to BIRD's value-grounded databases \citep{bird} and Spider~2.0's enterprise workflows \citep{spider2}; multilingual extensions include CSpider (Chinese) \citep{cspider}, AraSpider (Arabic) \citep{araspider}, and MultiSpider/MultiSpider~2.0 \citep{multispider,multispider2}. These resources establish that language-specific evaluation matters, but their central strategy is to translate the original Spider distribution. IndicDBBench is complementary: it creates a new evaluation environment from real Indian public data -- administrative hierarchies, agricultural and public-health concepts, and high-cardinality values -- testing whether models can ground Indic questions in the data systems Indian users actually encounter.

\textbf{Construction Methodology and External Validation.} Automatic-draft-then-human-verify is the field-standard recipe for multilingual Text-to-SQL benchmarks -- MultiSpider post-edits 37.1\% of NMT-drafted questions with three annotators per language (92.7\% agreement) \citep{multispider}, AraSpider revises ChatGPT drafts with bilingual post-editors \citep{araspider}, and CSpider translates questions over English schemas \citep{cspider} -- and machine-translated evaluation data is accepted beyond Text-to-SQL \citep{xnli, xtreme}, including Indic benchmarks built wholly or partly from English sources such as IndicGenBench and IndicXTREME \citep{indicgenbench, indicxtreme}. IndicDBBench follows the same pattern, prioritizing human verification by a reference-free quality-estimation score (Section~\ref{sec:qa}). We do not treat LLM translation as ground truth: dedicated NMT systems still outperform general-purpose LLM translators on Indic languages \citep{megaverse, indicgenbench}, which is why the screening-plus-audit protocol exists.

\textbf{Culturally Grounded and Indic Evaluation.} Linguistic coverage alone is insufficient when an evaluation set does not reflect the target setting: Falcon diagnoses schema, language mismatches in Chinese enterprise Text-to-SQL \citep{falcon}; IndicDBBench shares this motivation but targets the administrative, geographic, and socioeconomic structure of Indian open data. More broadly, IndicNLPSuite/IndicGLUE \citep{indicglue}, IndicXTREME \citep{indicxtreme}, IndicMT Eval \citep{indicmteval}, IndicGenBench \citep{indicgenbench}, and Indic-QA \citep{indicqa} establish that models and evaluation methods behave differently across Indic languages. Across this ecosystem, however, no benchmark evaluates grounded semantic parsing -- translating Indic-language questions into executable queries over real structured data. IndicDBBench extends the suite along this axis, combining parallel language variants with real relational databases and fine-grained failure analysis.

\section{The IndicDBBench Benchmark Construction}

IndicDBBench is built in three phases: \textbf{[1] Schema Synthesis} (flat government CSVs $\rightarrow$ rich relational structures), \textbf{[2] Task Generation} (value-grounded Text-to-SQL tasks), and \textbf{[3] Multilingual Expansion} (faithful Indic variants).

\begin{figure}[b]
  \begin{center}
    \includegraphics[width=\columnwidth]{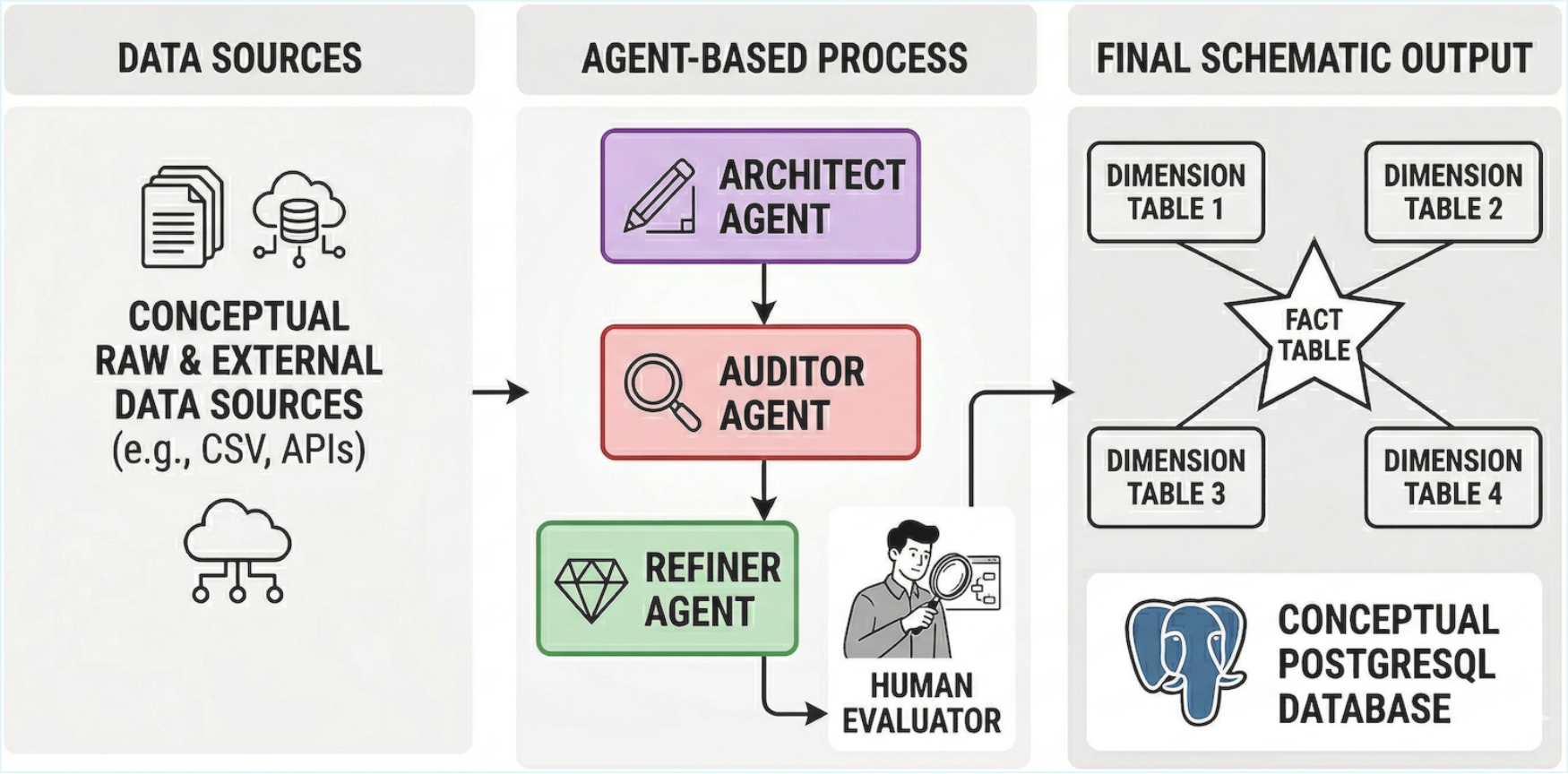}
    \caption{
      Pipeline for database schema generation.
    }
    \label{fig:agentic-pipeline}
  \end{center}
\end{figure}

\subsection{Agentic Schema Synthesis}
\label{sec:schema}

Indian open-data sources (NDAP, India Data Portal) distribute datasets as monolithic
CSVs with 50--100+ mixed-granularity columns. We convert these into complex relational
structures via a \textbf{3-Agent Judge Pattern} (as shown in Figure \ref{fig:agentic-pipeline}) - an iterative, LLM-driven feedback loop (see prompt \ref{fig:schema-gen-prompts}):

\begin{itemize}

\item \textbf{Architect} decomposes high-dimensional datasets into four to ten thematic entities, forming normalized star/snowflake schemas of Fact (FACT\_) and Dimension (DIM\_) tables capped at fifteen columns each. Quantitative metrics are centralized in Fact Tables and linked to surrounding Dimension Tables providing temporal or geographic context, ensuring that models must navigate multi-hop joins and precise schema linking.

\item \textbf{Auditor} validates the architecture against Third Normal Form and thematic cohesion, and checks relational-graph complexity so that key linkages necessitate multi-hop joins over at least three tables.

\item \textbf{Refiner} adjudicates between Architect and Auditor via an LLM-as-a-judge paradigm, standardizes column headers into canonical SQL identifiers, enforces strict typing, and emits a configuration file with a precise one-to-one mapping back to the denormalized source.

\end{itemize}

\noindent The agentic output compiles directly into a DDL file establishing foreign keys and surrogate keys for the administrative hierarchy (Country $\rightarrow$ State $\rightarrow$ District $\rightarrow$ Sub-District $\rightarrow$ Block $\rightarrow$ Village); after manual verification by database experts, it was executed and bulk-loaded into PostgreSQL (Appendix~\ref{app:schema-example} shows a generated schema).

\subsection{Task Synthesis via Enhanced DSQG-Syn}
\label{sec:tasks}

\begin{table}[t]
\centering
\small

\begin{tabular}{l l c c}
\toprule
\textbf{Category} & \textbf{Metric} & \textbf{Count} & \textbf{Pct. / Avg.} \\
\midrule
Volume 
 & Total Size & 15,612 & -- \\
 & Unique Pairs & 3,679 & -- \\

\midrule
Language 
 & English & 3,679 & 30.1 w \\
 & Hindi & 1,948 & 33.0 w \\
 & Indic-4* & 8,248 & 24.1 w \\
 & Hinglish & 1,737 & 29.0 w \\

\midrule
Difficulty 
 & Easy & 1,055 & 28.7\% \\
 & Medium & 1,539 & 41.8\% \\
 & Hard & 1,085 & 29.5\% \\

\midrule
SQL Op. 
 & JOIN & 3,484 & 94.7\% \\
 & WHERE & 3,278 & 89.1\% \\
 & GROUP BY & 2,441 & 66.3\% \\
 & ORDER BY & 2,289 & 62.2\% \\

\midrule
Agg. 
 & COUNT() & 929 & 25.3\% \\
 & SUM() & 809 & 22.0\% \\
 & AVG() & 560 & 15.2\% \\
\bottomrule
\end{tabular}

\vspace{2pt}
\raggedright \scriptsize *Indic-4 includes: Marathi, Bengali, Tamil, and Telugu.
\caption{Full quantitative statistics of the IndicDBBench benchmark.}
\label{tab:indicdb_full_stats}
\vskip -0.1in
\end{table}

\begin{table*}[t]
\centering
\small

\begin{tabular}{p{3.5cm} p{3.2cm} p{3.2cm} p{4cm}}
\toprule
\textbf{Feature / Constraint} & \textbf{Easy} & \textbf{Medium} & \textbf{Hard} \\
\midrule
Relational Depth & 0--1 JOIN & Exactly 1 JOIN & $\geq$ 2 JOINs \\
JOIN Diversity & INNER JOIN only & INNER JOIN primarily & Diverse (INNER, LEFT, RIGHT) \\
Filtering Logic & Simple \texttt{WHERE} & Moderate (e.g., ranges) & Complex multi-column filters \\
Aggregation & None & $\leq$ 1 clause & Required (GROUP BY) \\
Nesting & Prohibited & Prohibited & Required (CTEs/Sub-Q) \\
SQL Tokens & $<$ 60 & 60--120 & $>$ 120 \\
\bottomrule
\end{tabular}
\caption{Qualitative SQL complexity taxonomy used in IndicDBBench. Appendix~\ref{app:complexity} compares it with other benchmarks.}
\label{tab:indicdb_taxonomy}

\vskip -0.1in
\end{table*}

We adopt DSQG-Syn~\citep{dsqgsyn} for its \textbf{Question-First} paradigm: rather than randomly sampling columns (which yields intent-inconsistent pairs), it first generates domain-relevant questions across nine types covering all major SQL operations, then synthesizes grounded SQL--NLQ pairs. Our enhanced pipeline has four stages per database. \textbf{[1] Question generation:} BFS over the FK graph selects connected table subsets, and nine question types are generated per subset from LLM-extracted domain keywords. \textbf{[2] Schema linking:} a MAC-SQL-inspired selector \citep{Wang2025MACSQL} picks the minimal sub-schema, with sample values from PostgreSQL. \textbf{[3] Skeleton-guided SQL:} templates at three difficulty tiers (Easy 30\% / Medium 40\% / Hard 30\%; Table~\ref{tab:indicdb_taxonomy}) are filled with real schema names and \emph{real database values}, so filter values are never fabricated. \textbf{[4] NLQ synthesis:} questions deliberately avoid schema identifiers (``\textit{How many private clinics are there?}''), so parsers must show domain understanding rather than keyword matching.

\noindent \textbf{FK-Constrained Join Enforcement.}
The generator may follow only declared foreign-key paths -- preventing invalid joins such as \texttt{STATE\_ID}--\texttt{STATION\_ID} -- by injecting allowed relationships into the prompt and type-filtering numeric operations on non-numeric columns (prompt \ref{fig:dsqg-syn-prompts}).\\\\
\noindent \textbf{Task Statistics.}
The English dataset contains 3,679 validated NLQ--SQL pairs with a calibrated difficulty distribution (Table \ref{tab:indicdb_full_stats}). Integrity rests on two-tier validation -- PostgreSQL execution plus a manual audit by three database experts -- whose independent judgments reach Fleiss' $\kappa = 0.84$, validating the English question--SQL pairs.

\subsection{Multilingual Expansion}
\label{sec:multilingual}

\begin{figure*}[t]
  \begin{center}
    \centerline{\includegraphics[width=0.88\textwidth]{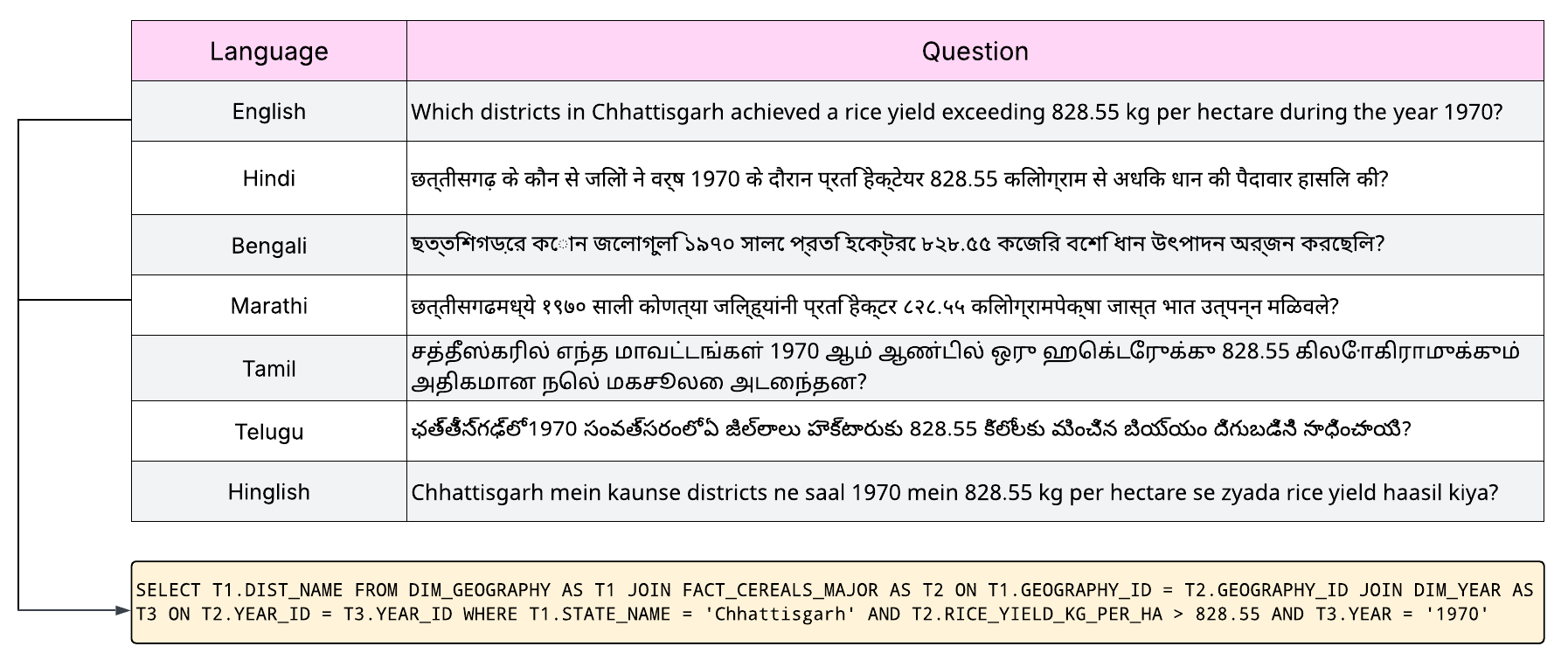}}
    \caption{
      Example of a generated multilingual task.
    }
    \label{fig:question-language-pair}
  \end{center}
  \vskip -0.2in
\end{figure*}

We expand English tasks into up to six additional variants -- \textbf{Hindi, Bengali, Tamil, Telugu, Marathi}, and \textbf{Hinglish} (HI--EN code-switching) -- yielding \textbf{15,612 task instances} (per-language coverage in Table~\ref{tab:indicdb_full_stats}). Only the NLQ is translated (English-First); the SQL remains identical, ensuring logical alignment across variants. Gemini 2.0 Flash is the primary conversion engine (Figure \ref{fig:question-language-pair}).\\\\
\textbf{Hinglish} receives specialized prompting for \emph{Natural Hinglish} -- Roman script blending Hindi grammar with English technical terms -- testing high-usage but low-resource patterns (prompt \ref{fig:translation-prompts}). A token-level audit of the generated Hinglish tasks (932 questions) confirms a balanced code-mixing ratio of \textbf{47.3\% English to 52.7\% Hindi} tokens.
\\

\noindent \textbf{Quality Assurance and Verification.}
\label{sec:qa}
We define a translated question as \emph{correct} if it is semantically equivalent to its English source and preserves all SQL-relevant content -- entities, literal filter values (translated into the target language and checked by the native-speaker reviewers), aggregation intent, and ordering direction -- such that a competent native speaker would map it to the same gold SQL. The reference for every variant is its English NLQ paired with the execution-validated gold SQL; automatic metrics only prioritize the human audit, never serving as ground truth. All translations follow a fixed rulebook governing schema tokens, database literals, administrative terminology, and named entities (Appendix~\ref{app:translation_protocol}). Verification proceeds in three stages:
\\\\
\noindent \textbf{Stage 1: Automated Screening.}
We score a stratified sample of 6,092 translations (1,032 per Indic language plus 932 Hinglish questions) with the reference-free \textbf{Unbabel/wmt22-cometkiwi-da} model \citep{cometkiwi}, the metric family best correlated with human MQM judgments for Indian languages \citep{indicmteval}. The 5,160 Indic translations reach $\mu = 0.820$, $\sigma = 0.0834$; Hinglish, scored but not thresholded, averages $0.669$ (Appendix~\ref{fig:comet-figs}). We corroborate this screen with two further automatic checks (Appendix~\ref{app:external-validation}): a second QE metric (\textbf{Unbabel/wmt23-cometkiwi-da-xl}) reproduces the same ranking (Kendall's $\tau$-b $0.46$--$0.57$) and detects the same 1,380 rule-derived hard failures (AUC $0.59$--$0.89$), and an IndicTrans2 \citep{indictrans2} back-translation round trip (chrF++, COMET-22 \citep{comet22}) matches or exceeds an IN22-Gen human-reference baseline in every language except Telugu. Because our templated questions are more repetitive than IN22-Gen's prose, we read this as a floor on gross semantic breakage, not as parity with human translation.
\\\\
\noindent \textbf{Stage 2: Statistical Thresholding.}
We set $\tau = \mu - 1\sigma = 0.737$ for targeted review: the 408 scored Indic translations below $\tau$ (7.9\%) are classified high-risk and routed to human review, concentrating expert auditing on the lower tail where semantic drift is most likely.
\\

\noindent \textbf{Stage 3: Native-Speaker Review.}
Every flagged translation, plus 200 randomly sampled unflagged translations per language (and 200 Hinglish questions, which are scored but not thresholded), was judged independently by two native speakers of that language against the correctness criterion above: \emph{Same} or \emph{Different}. A third native speaker resolved disagreements, and translations judged \emph{Different} were discarded.{ Raw agreement between the two reviewers is 94.8\% (Hindi), 95.3\% (Bengali), 91.0\% (Marathi), 94.0\% (Tamil), 92.4\% (Telugu), and 98.0\% (Hinglish), with Cohen's $\kappa$ of 0.67--0.77; 6.3\% of items needed a third-reviewer decision. Appendix~\ref{app:native-review} gives the full analysis on these labels, including validation of the COMET-QE threshold and an estimate of residual error among unflagged translations.} Reviewer profiles are in Appendix~\ref{app:translation_protocol}. Two error types recur among rejected translations, which we align with the MQM typology \citep{indicmteval} in Appendix~\ref{app:translation_protocol}:

\begin{itemize}
    \item \textbf{Lexical Entity Divergence} (MQM Accuracy/Mistranslation): mistranslation of domain-specific entities, such as agricultural varieties or administrative designations, which corrupts SQL \texttt{WHERE}-clause filters.
    \item \textbf{Logical Directional Inversion} (MQM Accuracy/Mistranslation, critical severity): sorting directives swapped in the target language (e.g., descending rendered as ascending), inverting the intended \texttt{ORDER BY} logic.
\end{itemize}

\begin{figure*}[t]
  \begin{center}
    \centerline{\includegraphics[width=0.88\textwidth]{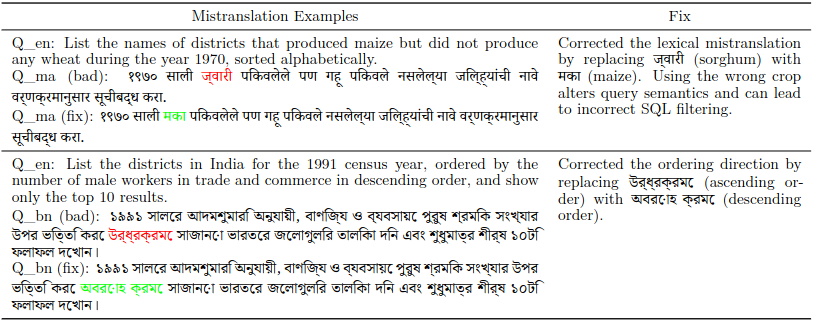}}
     \caption{
      Mistranslation examples found in the Stage-3 review. Each shows the English question (Q\_en), the flagged translation (bad; error in red), and a correct rendering for reference (fix; in green); the right column explains each error in English. Translations judged \emph{Different} were discarded, not corrected.
    }
    \label{fig:mistranslation_fixes}
  \end{center}
  \vskip -0.2in
\end{figure*}

\noindent Beyond these categories (examples in Figure \ref{fig:mistranslation_fixes}), rule-based checks also identified \textit{prompt leakage} -- English instructions retained in the Indic output. The observed gap persists across COMET-score strata (Appendix~\ref{app:translation-noise}). Because COMET is an imperfect proxy for semantic equivalence, these analyses do not exclude a contribution from translation errors. Marathi's higher flagging rate (26.0\% vs.\ 1.5--6.2\%) comes from how COMET-QE scores Marathi, not from worse translations, since its gap is the same on the COMET-screen-passing subset. Across all languages, the gap is 12.2 points on the best-translated third of questions and 11.1 on the worst.

\section{Experiments}

\subsection{Experimental Setup}
\label{sec:setup}

\textbf{Language Selection.~~}
Our selection follows a coverage-first principle. We evaluate seven variants: English (the baseline); \textbf{Hindi, Bengali, Marathi, Telugu, and Tamil} -- the five most widely spoken scheduled languages of India by first-language speakers (2011 Census), covering the Indo-Aryan and Dravidian families and four scripts, with a same-script pair (Hindi/Marathi) and a same-family, different-script pair (Tamil/Telugu); and \textbf{Hinglish}, the most prevalent code-mixed register in Indian usage. The binding constraint on coverage is validation rather than translation: our quality-assurance protocol requires bilingual experts per language (Section~\ref{sec:qa}).
To ensure a controlled comparison across languages, we keep the underlying database schema fixed and vary only the natural language queries via translation. We keep table and column names in English because the source platforms (NDAP and IDP) publish them in English, so real Indic-language users face the same mismatch. This mismatch alone does not explain the gap. Questions in six other non-English languages, asked against the same English schema, score close to English when evidence is provided (Section~\ref{sec:control_study}). Telling the model exactly which tables and values to use improves all languages by similar amounts and leaves a large estimated gap (Section~\ref{sec:main_results}).\\\\
\textbf{Models.~~} 
We evaluate models spanning architectures and scales: Llama 3.3 70B Instruct \citep{llama3} and Qwen3 8B \citep{qwen3} (decoder-only); DeepSeek V3.2 \citep{deepseek2025}, a mixture-of-experts model exceeding 671B total parameters; and MiniMax M2.7~\citep{minimax2026}, an agent-oriented model with self-evolving training. We additionally evaluate Sarvam 105B, Gemma 3 27B, GPT-5.6 Luna, and GPT-5.4 Mini on a 500-task stratified sample; the gap persists for all four (Appendix~\ref{app:extra-models}).\\\\
\textbf{Prompting Strategies.~~} 
We evaluate two prompting strategies: \textbf{zero-shot and DIN-SQL}~\citep{dinsql}, which decomposes generation into \textbf{[1]} schema linking, \textbf{[2]} clause-wise SQL construction, and \textbf{[3]} iterative self-correction (prompts in \ref{fig:prompt-online-synthesis}, \ref{fig:oneshot_system}). Zero-shot is run with and without evidence files generated per language with SEED~\citep{seed} (Appendix~\ref{app:evidence-generation}) from the question and schema only, never the gold SQL; they supply schema-linking cues, column values, and SQL hints. DIN-SQL is run with evidence.

To ensure experimental parity, we use identical prompt templates and in-context examples for all languages, with deterministic decoding (\textbf{temperature 0, top-p 1}).\\\\
\textbf{Evaluation metrics~~} 
We report Execution Accuracy (EX) \citep{spider, bird}: the share of tasks whose predicted query returns the same result set as the gold query.

\section{Experimental Results and Analysis}

\subsection{Main Results}
\label{sec:main_results}

\begin{table*}[t]
\centering
\small

\setlength{\tabcolsep}{3pt}%

\renewcommand{\arraystretch}{0.94}
\begin{tabular}{l l c c c c c c c}
\toprule
\textbf{Method} & \textbf{Model} & \textbf{English} & \textbf{Hindi} & \textbf{Bengali} & \textbf{Marathi} & \textbf{Tamil} & \textbf{Telugu} & \textbf{Hinglish} \\
\midrule

\multirow{4}{*}{DIN-SQL (w/ evidence)} 
& Llama 3.3 70B & 66.1\% & 58.0\% & 62.2\% & 54.5\% & 58.1\% & 58.0\% & 65.1\% \\
& Qwen3 8B               & 55.1\% & 52.7\% & 51.1\% & 49.4\% & 49.4\% & 50.0\% & 51.1\% \\
& DeepSeek V3.2          & \textbf{69.1\%} & \textbf{64.5\%} & \textbf{65.4\%} & 63.6\% & 66.7\% & \textbf{63.6\%} & \textbf{66.2\%} \\
& MiniMax M2.7            & 62.9\% & 59.1\% & 62.7\% & \textbf{63.7\%} & \textbf{68.1\%} & 62.0\% & 65.1\% \\

\midrule

\multirow{4}{*}{Zero-shot (w/o evidence)} 
& Llama 3.3 70B & 58.1\% & 44.2\% & 45.1\% & 42.2\% & 42.3\% & 39.5\% & 43.1\% \\
& Qwen3 8B               & 52.2\% & 38.4\% & 37.5\% & 36.4\% & 34.2\% & 34.5\% & 38.6\% \\
& DeepSeek V3.2          & \textbf{69.3\%} & \textbf{57.7\%} & \textbf{58.1\%} & \textbf{56.7\%} & \textbf{56.0\%} & \textbf{52.9\%} & \textbf{60.5\%} \\
& MiniMax M2.7            & 59.5\% & 53.3\% & 51.0\% & 52.0\% & 49.4\% & 48.2\% & 50.2\% \\

\midrule

\multirow{4}{*}{Zero-shot (w/ evidence)} 
& Llama 3.3 70B & 73.3\% & 61.3\% & 58.5\% & 57.1\% & 62.3\% & 47.7\% & 63.7\% \\
& Qwen3 8B               & 58.0\% & 42.4\% & 57.5\% & 53.0\% & 58.1\% & 45.5\% & 54.0\% \\
& DeepSeek V3.2          & 74.9\% & \textbf{67.7\%} & 66.7\% & \textbf{64.1\%} & \textbf{66.5\%} & \textbf{61.7\%} & \textbf{70.9\%} \\
& MiniMax M2.7            & \textbf{76.9\%} & 67.1\% & \textbf{67.1\%} & 63.3\% & 64.2\% & 57.7\% & 70.8\% \\

\midrule

\textbf{Max Performance Drop} &  & 0.0 & -11.4 & -11.8 & -13.0 & -14.3 & \textbf{-17.7} & -11.7 \\
\textbf{Avg. Drop per Language} &  & 0.0 & -9.1 & -7.7 & -9.9 & -8.3 & \textbf{-12.8} & -6.3 \\

\bottomrule
\end{tabular}
\caption{Execution accuracy (EX) across languages for different models and prompting strategies. \emph{Avg.\ Drop per Language} averages the twelve model--strategy rows; \emph{Max Performance Drop} is the largest drop vs.\ English among the three prompting strategies, each averaged over the four models. Drops are in percentage points. All results use the same 1,737 question--SQL pairs in each of the seven variants (12,159 tasks), selected from task IDs shared by every variant, so cross-language comparisons are matched.}
\label{tab:full_results}
\vskip -0.1in
\end{table*}

Table \ref{tab:full_results} presents the main results: across $12{,}159$ tasks and seven variants, a $9.0$-point global average drop vs.\ English -- consistent cross-lingual degradation.

\begin{figure}[t]
  \begin{center}
    \includegraphics[width=0.95\columnwidth]{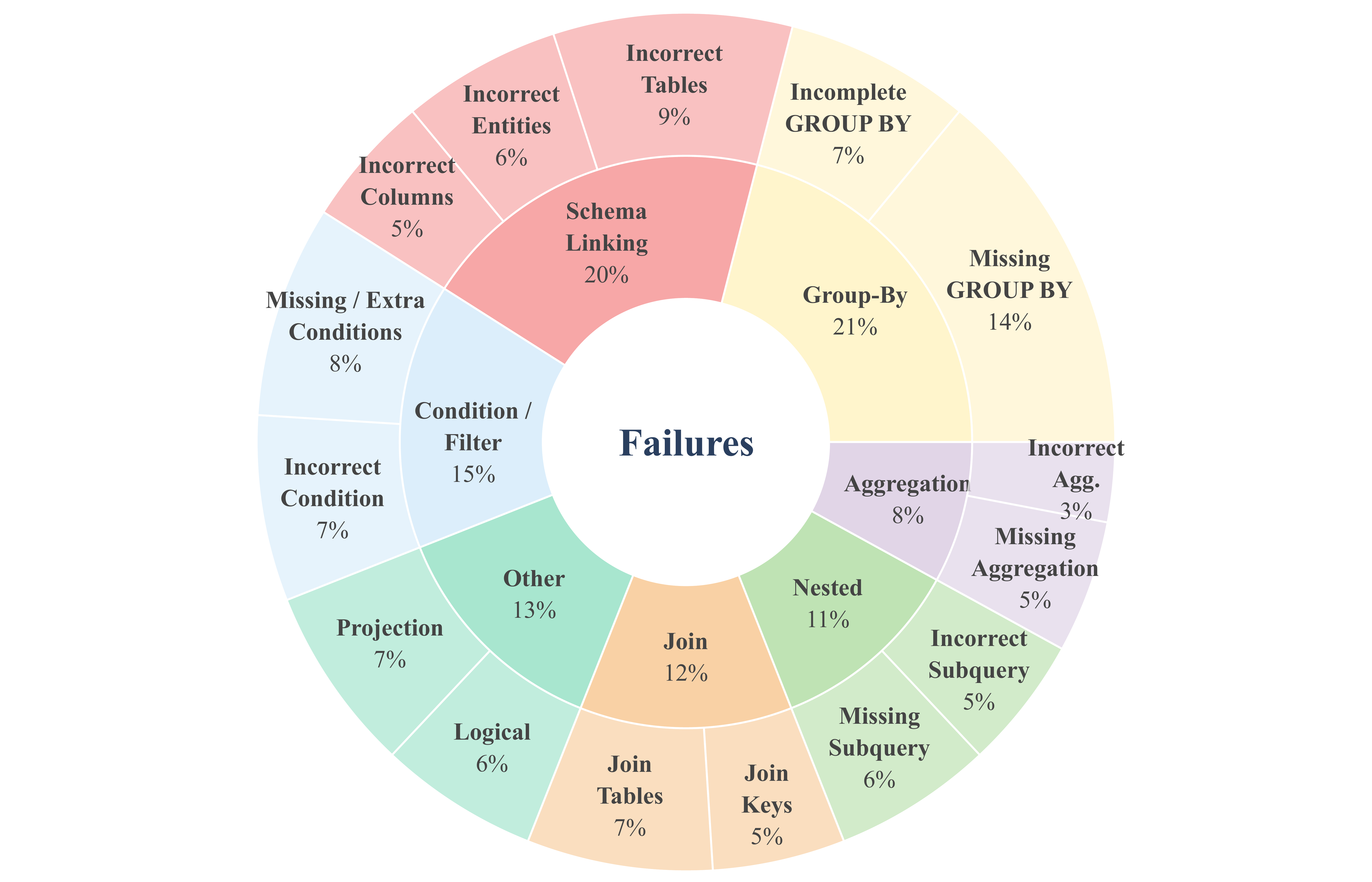}
    \caption{
      Distribution of error categories on the benchmark.
    }
    \label{fig:error_distribution}
  \end{center}
  \vskip -0.2in
\end{figure}

Averaged across strategies and models, degradation is consistent but uneven: Telugu exhibits the largest drop ($-12.8$ points) and Hinglish the smallest ($-6.3$), with Bengali, Tamil, Hindi, and Marathi between $-7.7$ and $-9.9$. To understand the causes, we analyze model errors across all languages; two categories dominate: schema linking and aggregation/group-by errors (Figure \ref{fig:error_distribution}).

\textbf{Schema linking errors} (20\%) misalign natural-language mentions with database elements -- tables, columns, and entities. These errors peak in Telugu, which exhibits the largest average decline (12.8 points). Quantified directly: the same question needs on average 6$\times$ more tokens in Telugu than English under the evaluated models' tokenizers, and tokenizer fertility correlates with the per-language drop both across languages ($r = +0.70$) and across all 24 model--language pairs ($r = +0.59$; Appendix~\ref{app:linguistic-analysis}); A controlled reordering experiment rules out subject--object--verb (SOV) word order as a driver (Appendix~\ref{app:sov}). This token--schema divergence complicates grounding; conversely, Hinglish shows the smallest drop, its English tokens aligning directly with schema elements.

\textbf{Does grounding explain the gap?} A controlled oracle ablation on Qwen3~8B and Llama~3.3~70B (the 500-task sample of Section~\ref{sec:control_study}; Appendix~\ref{app:oracle-decomp}) supplies gold \emph{tables} or gold \emph{entity values} on top of each question, holding all else fixed. Gold entity values target the value-presentation mismatch between Indic questions, which render literals in the target language, and English-valued databases; gold tables target table-level schema linking. Both signals raise accuracy in nearly every language, English included (the one exception is Qwen3 Telugu under gold tables, $-0.4$ points), so neither closes the gap: for Qwen3 it moves from 12.3 points to 10.0 with gold entities and 10.6 with both signals, for Llama from 11.0 to 9.4 and 11.4, and no change is significant under a task-level bootstrap (Table~\ref{tab:oracle-gap}). Grounding is still a strong lever for Indic users: both signals together raise mean Indic EX by 7.8 points for Qwen3 and 10.2 for Llama. But English rises by about as much, so a large estimated gap remains for both models. The oracle does not cover column selection or join keys, and its intervals are wide (for Qwen3 oracle entities, $[-6.3, +1.7]$ points against a 12.3-point gap), so the study does not determine how much broader grounding would explain.

\textbf{Aggregation and group-by errors} (29\%) are the largest structural category -- missing or incomplete GROUP BY clauses and incorrect aggregation -- reflecting compositional-reasoning limits that multilingual variation amplifies: shifts in how quantitative intent is expressed obscure query structure, so models capture the right entities but build the wrong operations (case studies in Appendix~\ref{sec:case_studies}).

\textbf{Stratified analyses.} We slice the task-matched gap by difficulty tier and gold-query structure. Under zero-shot prompting the relative gap widens with difficulty, from 15.3\% of the English score on Easy tasks to 22.5\% on Hard (Telugu doubles, 15\%$\rightarrow$31\%); DIN-SQL's schema linking and decomposition shrink it to $\sim$4 points and flatten it across tiers, isolating a difficulty-independent schema-linking floor, largest for Marathi (Appendix~\ref{app:difficulty-gap}). The gap is also larger on multi-join, grouping, and multi-hop queries than on simple ones (Table~\ref{tab:structural-gap}).

\subsection{Ablation Study: Use of Evidence}
\label{sec:ablation}

Under DIN-SQL, on a random half (6,080 tasks) of Table~\ref{tab:full_results}'s 12,159-task evaluation set, \textit{DeepSeek V3.2} gains +22.9 to +29.0 EX points from structured evidence (Table~\ref{tab:evidence_ablation}). Appendix~\ref{sec:ablation_statistics} traces these gains to improved grounding between question phrases, English schema elements, and database values. Relative to English (+25.6), gains are larger for Tamil (+29.0) and Marathi (+28.0), about equal for Telugu and Bengali, and smaller for Hindi (+24.0) and Hinglish (+22.9), so for DeepSeek V3.2 evidence narrows no gap by more than 3.4 points and widens two. Across the zero-shot rows of Table~\ref{tab:full_results}, evidence narrows the gap in most model--language pairs, but by amounts that vary widely by model. An evidence-language ablation (Appendix~\ref{app:evidence-generation}) shows comparable gains whether the evidence is written in English or the target language: the benefit lies in the grounding and partial planning it encodes, not in English as its medium.

\begin{table}[b]
\centering
\small

\setlength{\tabcolsep}{3pt}

\renewcommand{\arraystretch}{0.94}
\begin{tabular}{lccc}
\toprule
\textbf{Language} & \textbf{Without evidence} & \textbf{With evidence} & \textbf{$\Delta$ (pp)} \\
\midrule
English  & 45.0\% & 70.6\% & +25.6 \\
Hindi    & 39.8\% & 63.8\% & +24.0 \\
Bengali  & 40.6\% & 66.2\% & +25.6 \\
Marathi  & 36.8\% & 64.8\% & +28.0 \\
Tamil    & 39.8\% & 68.8\% & +29.0 \\
Telugu   & 37.9\% & 63.6\% & +25.7 \\
Hinglish   & 41.9\% & 64.8\% & +22.9 \\
\bottomrule
\end{tabular}
\caption{Execution accuracy (EX) of DeepSeek V3.2 under DIN-SQL with and without evidence file augmentation across languages.}
\label{tab:evidence_ablation}
\vskip -0.1in
\end{table}

\subsection{Is the Indic Gap Specific to Indic Languages?}
\label{sec:control_study}

Because IndicDBBench keeps schema identifiers, SQL, and many values in English, the Indic Gap could reflect any non-English input or script mismatch. To separate these factors, we translate the evaluation sample into six non-Indic \emph{control} languages -- Spanish, French, Italian (Latin); Arabic, Chinese, Korean (non-Latin) -- and evaluate the same four models under all three conditions on the oracle study's 500-task stratified sample (Appendix~\ref{sec:control_appendix}), so conditions differ only in question language.

\begin{table}[t]
\centering
\small
\setlength{\tabcolsep}{3pt}
\begin{tabular}{lrrrrr}
\toprule
\textbf{Condition} & \textbf{En} & \textbf{Lat} & \textbf{N-Lat} & \textbf{$\Delta$L} & \textbf{$\Delta$N} \\
\midrule
DIN-SQL (w/ ev.)      & 67.5 & 69.4 & 67.5 & $-1.9$ & $0.0$ \\
Zero-shot (w/o ev.)   & 61.7 & 58.2 & 49.4 & $+3.5$ & $+12.3$ \\
Zero-shot (w/ ev.)    & 63.7 & 65.4 & 61.5 & $-1.7$ & $+2.2$ \\
\bottomrule
\end{tabular}
\caption{Execution accuracy (EX, \%) averaged over four models on the 500-task sample. Lat = Spanish, French, Italian; N-Lat = Arabic, Chinese, Korean. $\Delta$ = drop vs.\ English (positive = English better). Per-model results in Appendix~\ref{sec:control_appendix}.}
\label{tab:control_groups}
\vskip -0.1in
\end{table}

Table~\ref{tab:control_groups} summarises the three conditions. With evidence, Latin-script controls sit $1.9$ and $1.7$ points \emph{above} English (DIN-SQL and zero-shot respectively); non-Latin controls are level under DIN-SQL ($67.5$ vs $67.5$) but $2.2$ points below English zero-shot. Without evidence both groups fall behind (non-Latin $-12.3$ points, Latin-script $-3.5$): script and tokenization mismatch surfaces mainly when the model must resolve schema references unaided. With evidence the controls stay far closer to English than the Indic variants, whose average gaps in the same conditions are $3.7$ (DIN-SQL) and $10.4$ points (zero-shot; Table~\ref{tab:full_results}). Without evidence the non-Latin gap ($12.3$) approaches the Indic one ($13.0$), although on a different task sample. Non-English input and non-Latin script therefore do not explain the gap that persists once grounding is provided. We attribute that residual to lower Indic pretraining representation combined with the cross-lingual schema-linking demands of Indian administrative terminology.

\section{Conclusion}
We presented IndicDBBench, a benchmark for cross-lingual semantic parsing over Indian administrative data: a three-agent judge pattern builds star/snowflake schemas across 237 tables, whose tasks are screened with COMET-QE and audited by native speakers on flagged and randomly sampled translations. We find a persistent 9.0-point Indic Gap. Oracle tables and literal values raise accuracy in every language by similar amounts and evidence narrows the gap only modestly; how much broader grounding, such as column and join selection, would explain remains open.

\section*{Limitations}
This study provides an initial exploration of large language model cross-lingual capabilities in the Indian context. Future releases will prioritize Kannada and Malayalam -- completing the scheduled Dravidian languages, with Kannada offering a script structurally close to Telugu but morphologically distinct, a natural controlled comparison for the family effects we observe -- followed by Gujarati, Punjabi, and Odia to broaden Indo-Aryan and script coverage. Because IndicDBBench is English-first, expansion requires only translation and validation of the natural-language questions: schemas, gold SQL, and evidence files are reused unchanged. High-quality open translation models covering all 22 scheduled languages \citep{indictrans2}, combined with our released COMET-QE screening pipeline, make community-driven extensions practical; expert human validation remains the deliberate bottleneck. The questions themselves are generated rather than collected from users. Two-tier validation -- PostgreSQL execution and a manual audit by three database experts (Fleiss' $\kappa = 0.84$, Section~\ref{sec:tasks}) -- establishes that each NLQ--SQL pair is correct and executable, but checking that every item is correct is not the same as showing that the set is representative: the nine question types, the FK-guided table subsets, and the 30/40/30 difficulty split are design choices fixed in advance, not distributions measured from a deployed system. The questions are accordingly more templated and lexically less varied than free-form user requests, a property visible in their back-translation behaviour (Table~\ref{tab:backtranslation}). What IndicDBBench grounds in reality is therefore the data rather than the question distribution: schemas are derived from Indian open-data platforms and predicate values are drawn from the databases themselves rather than fabricated. Whether the Indic Gap reported here holds under the query distribution of a deployed system remains open. Benchmarks drawn from naturally occurring usage -- Stack Exchange query logs in SEDE~\citep{sede}, enterprise warehouse workloads in BEAVER~\citep{beaver}, and questions elicited from 222 hospital staff in EHRSQL~\citep{ehrsql} -- indicate what such validation requires; to our knowledge no comparable resource is publicly available for an Indic-language database interface. Because IndicDBBench retains English table and column names while translating the natural-language questions, the observed Indic Gap may reflect not only differences in compositional SQL reasoning but also the difficulty of cross-lingual schema linking. The reduction in this gap under evidence augmentation (Table \ref{tab:evidence_ablation}) suggests that explicit grounding signals may alleviate part of this bottleneck, although oracle grounding of tables and literal values leaves a large estimated gap in place (Appendix~\ref{app:oracle-decomp}); the oracle does not cover column choice, so a localized-schema comparison remains needed to fully separate the two factors. The non-Indic control study (Section \ref{sec:control_study}) shows that neither non-English input nor non-Latin script accounts for the full gap, but it is run on a 500-task stratified sample with four models under the three prompting conditions of the main experiments, and it therefore establishes a matched comparison rather than a causal decomposition. Future work should further disentangle these factors by comparing English-schema settings with fully or partially localized schemas, including regional-language column names used in some Indian state-government databases. Our evaluation is restricted to prompting-based methods; benchmarking fine-tuned and agentic Text-to-SQL systems on IndicDBBench is an important next step, and there is significant potential for supervised fine-tuning and retrieval-augmented generation to address performance deficits in high-cardinality environments. A post-hoc quality sweep identified residual translation artifacts in a subset of items; task-matched analysis bounds their effect on the measured per-language gaps to within $\pm$1.3 points (Appendix~\ref{app:translation-noise}). All five Indic languages are verb-final (SOV); a controlled constituent-order experiment (Appendix~\ref{app:sov}) found no evidence that word order drives the gap -- across Llama 3.3, DeepSeek V3.2, and Qwen3, intent-fronting never significantly improved any Indic variant, while reordering significantly degraded DeepSeek V3.2 on Bengali and Telugu, indicating sensitivity to natural phrasing rather than to constituent order. Furthermore, future benchmark iterations will implement automated methods to mitigate logical inversions and lexical divergences identified during error analysis. Finally, further research is required to examine how multi-turn interactions and agentic workflows impact the reliability of multilingual Text-to-SQL synthesis across diverse relational frameworks.

\section*{Impact Statement}

This work presents a multilingual Text-to-SQL benchmark aimed at advancing research in cross-lingual semantic parsing. By enabling evaluation across diverse Indic languages, it has the potential to improve the accessibility of data systems for non-English speakers. Potential risks include the propagation of biases present in the underlying datasets and unequal performance across languages, which could disadvantage certain user groups. We release this benchmark to encourage the development of more inclusive and robust multilingual systems, and we hope it will facilitate further research into fairness and reliability in cross-lingual AI. At the camera-ready stage, we will release the complete databases, evidence files, prompts, and evaluation scripts to support reproducibility and future research.

\bibliography{custom}

\appendix
\section{Appendix}
\label{sec:appendix}

\subsection{Case Study Examples}
\label{sec:case_studies}
\label{fig:case-study}
\begin{figure*}[t]
\centering
\includegraphics[width=0.97\textwidth]{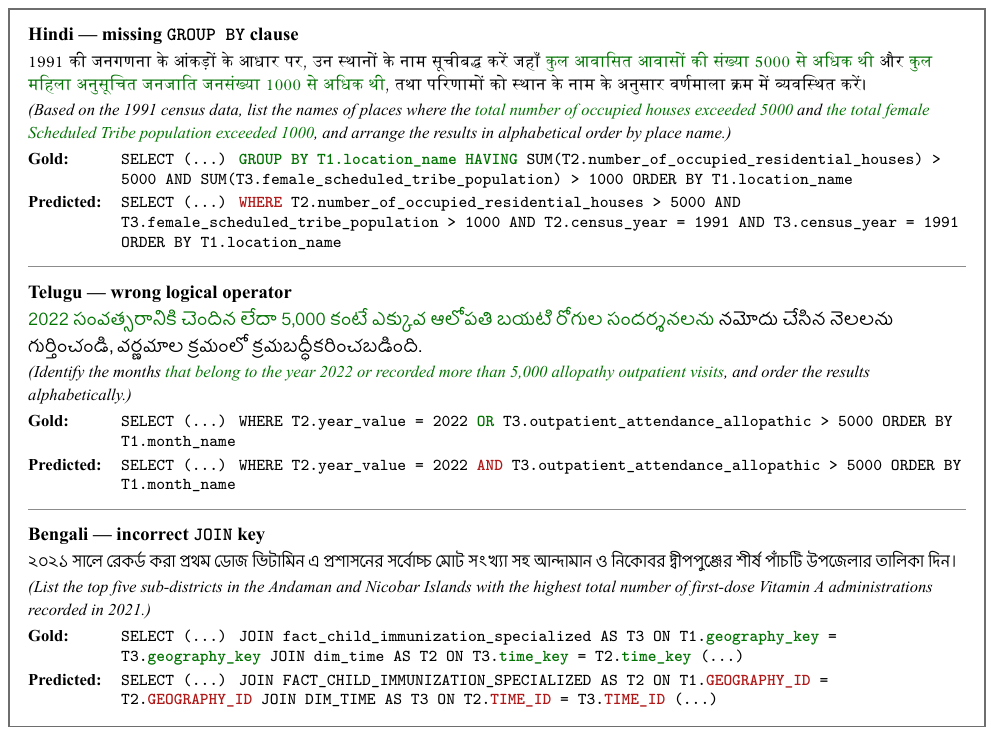}
\caption{Case studies of the three dominant error classes (cf.\ Figure~\ref{fig:error_distribution}), one per language. \textbf{Hindi:} the per-group thresholds (green) are compiled as row-level \texttt{WHERE} filters, dropping the required \texttt{GROUP BY}/\texttt{HAVING}. \textbf{Telugu:} an explicit disjunction (year 2022 \emph{or} more than 5{,}000 outpatient visits) is collapsed into \texttt{AND}. \textbf{Bengali:} the join path is correct, but the join keys are hallucinated \texttt{\_ID} identifiers in place of the schema's \texttt{\_key} columns. Each panel shows the Indic question and its English gloss (green spans mark the constraint phrases), the gold SQL with the decisive clause in green, and the model prediction with the faulty fragment in red.}
\label{fig:case_studies}
\end{figure*}

Figure~\ref{fig:case_studies} walks through one representative failure from each of the dominant error categories of Figure~\ref{fig:error_distribution}. In every case the English gloss shows that the Indic question faithfully preserves the source intent; the failures arise \emph{after} comprehension, in mapping a correctly understood question onto SQL structure.

\noindent \textbf{Hindi: missing \texttt{GROUP BY} (aggregation class, the largest at 29\%).} The question asks for places where the \emph{total} number of occupied houses exceeds 5{,}000 and the \emph{total} female Scheduled-Tribe population exceeds 1{,}000: both thresholds apply to per-place sums, which requires \texttt{GROUP BY location\_name} with \texttt{HAVING SUM(...)} conditions, as in the gold query. The model resolves every entity and column correctly but applies both thresholds as row-level \texttt{WHERE} filters, silently changing the question from ``places whose totals exceed the thresholds'' to ``places with any single record exceeding them.'' The aggregation intent is carried by one word (``kul'', \emph{total}), and once it is missed the rest of the query assembles plausibly but wrongly.

\noindent \textbf{Telugu: inverted logical connective (condition/logic class).} The question asks for months that belong to 2022 \emph{or} recorded more than 5{,}000 allopathic outpatient visits. The disjunction links two relative clauses rather than two adjacent predicates, and the model renders it as \texttt{AND}, collapsing the union into an intersection that returns only 2022 months with high attendance. Entities, thresholds, and ordering are all correct; only the connective's polarity is lost.

\noindent \textbf{Bengali: fabricated join keys (schema-linking class, 20\%).} The model selects the correct three tables for the Vitamin-A question but joins them on \texttt{GEOGRAPHY\_ID} and \texttt{TIME\_ID}, conventional-sounding identifiers that do not exist in this schema, whose actual keys are \texttt{geography\_key} and \texttt{time\_key}. This is grounding failure in its purest form: the model substitutes the naming convention it expects for the vocabulary the schema declares. It is precisely the failure mode that evidence files repair most often, with column selection and value mapping jointly accounting for over 60\% of evidence-driven corrections (Appendix~\ref{sec:ablation_statistics}).

\noindent Across all three cases the predictions are fluent, executable SQL with correctly resolved entities; what fails is structural (aggregation scope, connective polarity, key grounding) rather than translation, consistent with the task-matched analysis in Appendix~\ref{app:translation-noise}. This is also why structured decomposition helps: DIN-SQL's explicit schema-linking and clause-wise construction absorbs most of the difficulty-dependent component of these errors, leaving the flat schema-linking floor documented in Appendix~\ref{app:difficulty-gap}.

\subsection{SQL Complexity Relative to Existing Benchmarks}
\label{app:complexity}

\begin{table*}[t]
\centering
\small
\setlength{\tabcolsep}{5pt}
\begin{tabular*}{\textwidth}{@{\extracolsep{\fill}}llp{8.2cm}}
\toprule
\textbf{Benchmark} & \textbf{Tables/DB} & \textbf{Query structure, as reported} \\
\midrule
Spider        &  5.1  & 0.6 joins/query; 1.1 nesting/query \\
BIRD          &  6.8  & 0.9 joins/query; 1.1 nesting/query \\
IndicDBBench  & 11.85 & 94.7\% of queries with $\geq$1 JOIN, chains to depth six; 66.3\% GROUP BY \\
BEAVER        & 101.5 & 5.7 joins/query; 5.6 nesting/query \\
\bottomrule
\end{tabular*}
\caption{Structural comparison with existing Text-to-SQL benchmarks, ordered by schema size. Spider, BIRD, and BEAVER figures are those reported by \citet{beaver} (Table~3), which measures all three under one protocol; IndicDBBench figures are from Table~\ref{tab:indicdb_full_stats}. Each benchmark's statistics are reproduced as published rather than recomputed, so the final column is not on a common scale.}
\label{tab:complexity-comparison}
\end{table*}

IndicDBBench places difficulty on two axes at once: the questions are deliberately underspecified, so a parser must infer which schema elements are intended (Section~\ref{sec:tasks}), and the gold queries are structurally demanding in their own right. The second axis is the one most readily checked against existing resources.

On schema size, IndicDBBench averages 11.85 tables per database (237 tables across 20 databases), more than twice Spider and 1.7 times BIRD (Table~\ref{tab:complexity-comparison}). On query structure, 94.7\% of gold queries contain at least one JOIN, with join chains reaching depth six, and 66.3\% use GROUP BY (Table~\ref{tab:indicdb_full_stats}); Spider reports 26\% GROUP BY~\citep{spider}, and its mean of 0.6 joins per query bounds the share of Spider queries containing any JOIN at 60\%. IndicDBBench is therefore structurally heavier than the cross-domain datasets it is most often compared with, while remaining well short of the enterprise setting: Spider~2.0~\citep{spider2} and BEAVER~\citep{beaver} target workflow-scale SQL over far larger schemas, which IndicDBBench does not attempt. Extending coverage in that direction is planned work.

SQL token count is one of six criteria in the difficulty taxonomy of Table~\ref{tab:indicdb_taxonomy}, alongside relational depth, join diversity, filtering logic, aggregation, and nesting, and is not used on its own. Length is an established component of difficulty in this literature: BIRD reports tokens per query as evidence of query intricacy~\citep{bird}, Spider~2.0 assigns difficulty tiers by token count~\citep{spider2}, and \citet{archer} report execution accuracy falling as gold SQL grows longer.

\subsection{Translation Protocol and Annotator Profile}
\label{app:translation_protocol}

\noindent \textbf{Translation Rulebook.} All six non-English variants were produced under the following fixed rules, applied through the translation prompts (Appendix~\ref{fig:translation-prompts}) and enforced during human review:

\begin{enumerate}
    \item \textbf{Schema identifiers and SQL.} Table names, column names, and the gold SQL are never translated; the SQL is shared verbatim across all seven variants.
    \item \textbf{Administrative terminology.} Generic administrative levels are rendered with the target language's standard census-hierarchy term, e.g., \emph{zila}/\emph{upa-zila} (Hindi), \emph{tunai mavattam} (Tamil), \emph{upa-jilla} (Telugu), and \emph{upajilhe} (Marathi) for \texttt{SUB\_DISTRICT}, rather than region-specific units such as \emph{taluk}, \emph{mandal}, or \emph{panchayat}, because the underlying NDAP/IDP schemas encode the pan-India census hierarchy (State $\rightarrow$ District $\rightarrow$ Sub-District $\rightarrow$ Block $\rightarrow$ Village). This choice makes schema linking harder than retaining English administrative terms would, while avoiding conflation of regionally distinct units; terminology consistency was explicitly checked during the Stage-3 review (Section~\ref{sec:multilingual}).
    \item \textbf{Named entities.} Proper nouns are transliterated according to the accepted convention of the target language; new coinages are avoided, following the IN22 translation guidelines \citep{indictrans2}.
    \item \textbf{Numbers, dates, and units.} Left unchanged from the English source rather than localized to target-language numeral, date, or unit conventions.
    \item \textbf{Hinglish.} Roman script with a Hindi grammatical frame; all schema tokens and SQL-critical values remain in English, and the per-task code-mixing ratio is tracked in the released data.
\end{enumerate}

\noindent \textbf{Annotator Profile.} Following IndicGenBench \citep{indicgenbench} and IndicMT Eval \citep{indicmteval}, each language was reviewed by two native speakers, and a third native speaker resolved disagreements. All reviewers are fluent in English, were schooled in the target language, and have at least two years of English--Indic translation or annotation experience. Hinglish was reviewed by Hindi--English bilinguals who use Hinglish daily. English question--SQL pairs were checked separately by three database experts (Section~\ref{sec:tasks}).

\noindent \textbf{Error Taxonomy (MQM Alignment).} Table~\ref{tab:mqm_mapping} maps the error categories of Section~\ref{sec:multilingual} onto the Multidimensional Quality Metrics (MQM) typology used for Indic MT evaluation \citep{indicmteval}.

\begin{table}[h]
\centering
\small
\setlength{\tabcolsep}{3pt}
\begin{tabular}{p{0.27\linewidth} p{0.27\linewidth} p{0.36\linewidth}}
\toprule
\textbf{IndicDBBench category} & \textbf{MQM category} & \textbf{Severity / SQL impact} \\
\midrule
Lexical Entity Divergence & Accuracy / Mistranslation & Major: corrupts \texttt{WHERE}-clause filter values \\
Logical Directional Inversion & Accuracy / Mistranslation & Critical: inverts \texttt{ORDER BY} direction \\
Prompt Leakage & Non-translation / Source artifact & Major: extraneous instructions retained in the NLQ \\
\bottomrule
\end{tabular}
\caption{Mapping of IndicDBBench translation-error categories onto the MQM typology \citep{indicmteval}.}
\label{tab:mqm_mapping}
\end{table}

\subsection{Analyses Enabled by the Native-Speaker Review}
\label{app:native-review}

\noindent\textbf{Design.} The review packets contain 1,608 items: all 408 COMET-flagged Indic translations ($<\tau$), 200 unflagged translations sampled at random per Indic language, and 200 Hinglish questions. Items are shuffled, so reviewers cannot tell flagged items from random ones. Two native speakers label each item \emph{Same} or \emph{Different} independently. They also record an error type and a naturalness rating. A third native speaker resolves every disagreement.

\begin{table}[h]
\centering
\small
\setlength{\tabcolsep}{2.5pt}
\begin{tabular}{lrcc cc}
\toprule
 & & & & \multicolumn{2}{c}{\textbf{Final \emph{Different} (\%)}} \\
\cmidrule(lr){5-6}
\textbf{Lang.} & \textbf{Items} & \textbf{$\kappa$} & \textbf{Agr.} & \textbf{Random} & \textbf{Flagged} \\
\midrule
Hindi    & 229 & 0.67 & 94.8 & 4.0 {\scriptsize[2.0, 7.7]} & 37.9 {\scriptsize[22.7, 56.0]} \\
Bengali  & 215 & 0.68 & 95.3 & 5.0 {\scriptsize[2.7, 9.0]} & 40.0 {\scriptsize[19.8, 64.3]} \\
Marathi  & 468 & 0.75 & 91.0 & 4.0 {\scriptsize[2.0, 7.7]} & 37.3 {\scriptsize[31.7, 43.2]} \\
Tamil    & 232 & 0.69 & 94.0 & 6.0 {\scriptsize[3.5, 10.2]} & 37.5 {\scriptsize[22.9, 54.7]} \\
Telugu   & 264 & 0.72 & 92.4 & 8.0 {\scriptsize[5.0, 12.6]} & 39.1 {\scriptsize[28.1, 51.3]} \\
Hinglish & 200 & 0.77 & 98.0 & 4.0 {\scriptsize[2.0, 7.7]} & n.a. \\
\midrule
Pooled   & 1,608 & 0.74 & 93.7 & 5.2 {\scriptsize[4.1, 6.6]} & 37.7 {\scriptsize[33.2, 42.5]} \\
\bottomrule
\end{tabular}
\caption{Agreement between the two reviewers (Cohen's $\kappa$ and raw agreement on the initial binary labels), and the share of items judged \emph{Different} after adjudication, among randomly sampled unflagged items and among COMET-flagged items. Brackets give 95\% Wilson intervals. Hinglish is not thresholded, so it has no flagged group.}
\label{tab:native-review}
\end{table}

\noindent\textbf{What the review measures that automatic checks cannot.} On these labels, the protocol yields four results that no automatic metric in Section~\ref{sec:qa} can provide.

\begin{enumerate}
    \item \textbf{Reliability of the gold standard.} Per-language Cohen's $\kappa$ (0.67--0.77) and raw agreement (91.0--98.0\%) quantify how reproducible the \emph{Same}/\emph{Different} judgment is. This is the inter-annotator figure that the field-standard recipe reports \citep{multispider}. Adjudication by a third reviewer was needed for 6.3\% of items (102 of 1,608).
    \item \textbf{Validation of the COMET-QE screen.} Flagged items are judged \emph{Different} 7.3 times as often as random unflagged items (37.7\% vs.\ 5.2\%, pooled), and the two intervals do not overlap. This tests directly whether $\tau = 0.737$ concentrates real errors, rather than assuming it.
    \item \textbf{An unbiased estimate of residual error.} The random arm estimates the error rate among translations the screen passes. Weighting the two arms by each language's flagged share gives a corpus-level error estimate of 5.0\% (Hindi), 5.5\% (Bengali), 12.7\% (Marathi), 7.0\% (Tamil), and 9.9\% (Telugu), or 8.0\% pooled over the five Indic languages. In every language except Marathi, most estimated errors (76--89\%) fall \emph{above} $\tau$, where COMET-QE alone would not catch them. Only the random human sample reveals this.
    \item \textbf{Error type and naturalness.} Reviewers assign each \emph{Different} item one of seven SQL-relevant error types (wrong entity or filter value, number, aggregation, sort direction, missing or extra condition, untranslated text, other) and rate every item's naturalness. In the labels, the error types are spread almost evenly (12--15\% each), and 82.8\% of ratings are \emph{Natural} (78.7\% Marathi to 87.2\% Hinglish).
\end{enumerate}

\noindent Two further analyses use the same packets. The first recomputes the Indic Gap on the items that both reviewers judge \emph{Same}, which bounds the gap that remains after human-verified translation. The second compares the Marathi flagged and random arms, which tests the calibration explanation of Appendix~\ref{app:translation-noise} directly. 

\subsection{COMET-QE Scores}
\label{fig:comet-figs}

\begin{figure*}[t]
\centering
\begin{minipage}[t]{0.49\textwidth}
\centering
\includegraphics[width=\linewidth]{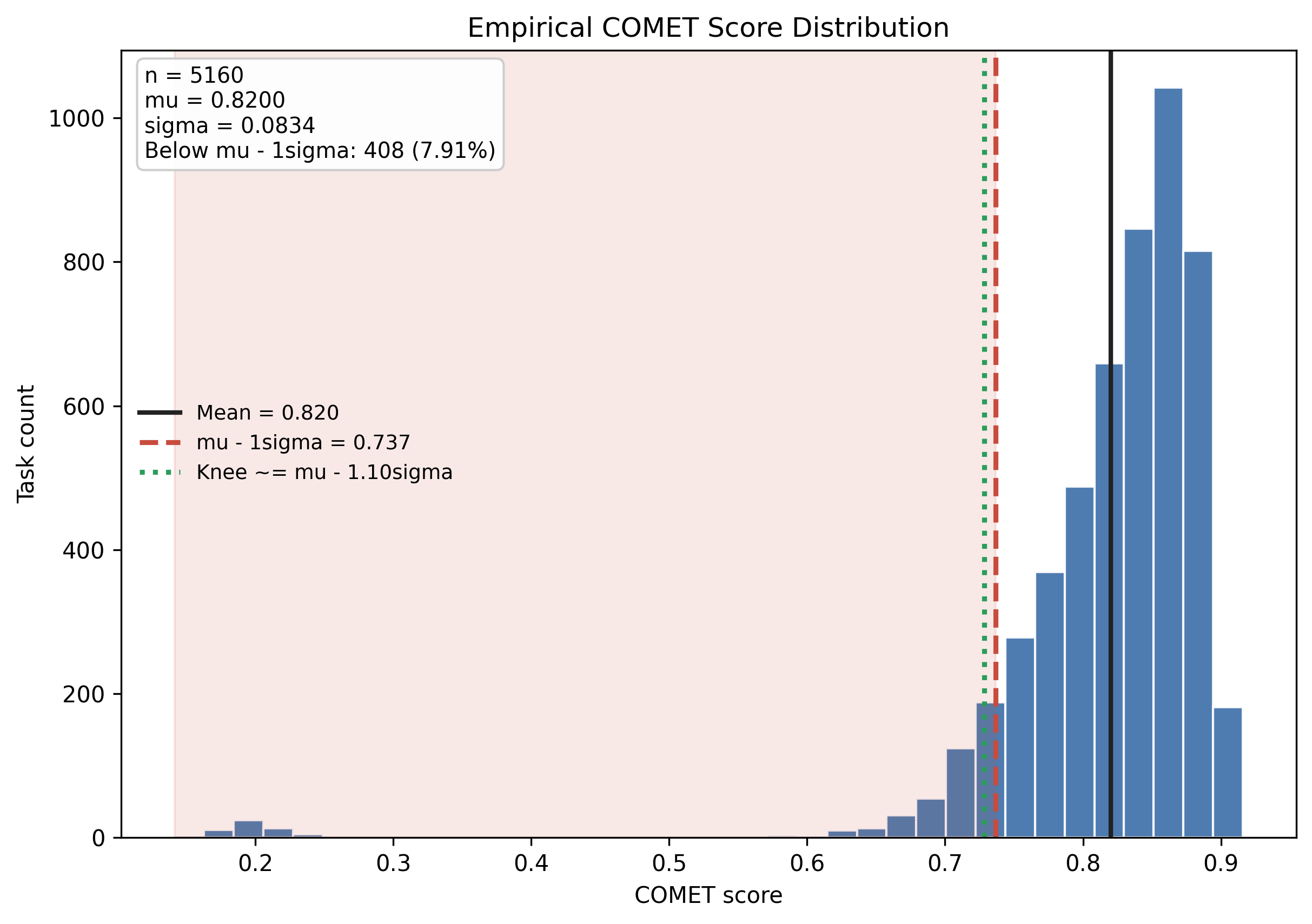}
\end{minipage}\hfill
\begin{minipage}[t]{0.49\textwidth}
\centering
\includegraphics[width=\linewidth]{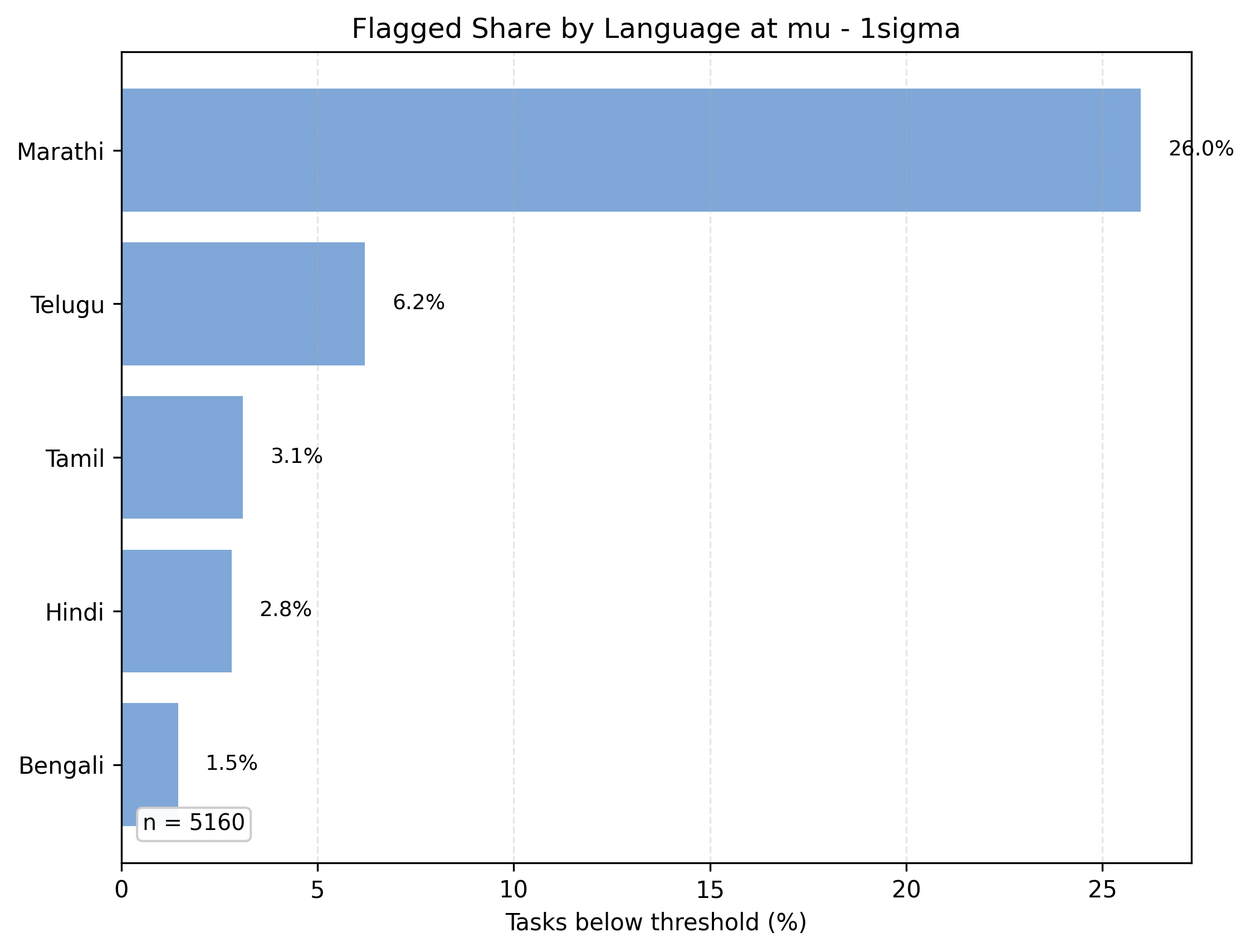}
\end{minipage}
\caption{COMET-QE quality score distributions: (Left) aggregate distribution across the corpus, (Right) language-specific breakdown detailing the variance used for targeted human audit.}
\label{fig:comet-distributions-combined}
\end{figure*}

Figure~\ref{fig:comet-distributions-combined} summarizes the automated screening stage of the quality-assurance pipeline (Section~\ref{sec:qa}). The left panel shows the empirical distribution of reference-free COMET-QE scores over the 5,160 screened Indic translations, the stratified sample of Section~\ref{sec:qa} ($\mu = 0.820$, $\sigma = 0.0834$); the marked threshold $\tau = \mu - 1\sigma = 0.737$, isolates the lower tail of the distribution, the 408 translations (7.9\% of the screened set) routed to expert review. Hinglish is excluded from thresholding because COMET-family metrics are unreliable on romanized code-mixed text; Hinglish variants were audited directly by bilingual reviewers.

The right panel breaks the flagged share down by language and shows a pronounced asymmetry: 26\% of Marathi translations are flagged at $\tau$ (two thirds of all flags), an order of magnitude above Bengali (1.5\%) and well above Telugu (6.2\%), Tamil (3.1\%), and Hindi (2.8\%). This asymmetry motivated the targeted analysis in Appendix~\ref{app:translation-noise}, which shows it reflects COMET-QE calibration for Marathi rather than translation damage: the Marathi--English gap is essentially unchanged on the COMET-screen-passing subset, and flagged Marathi items show no execution-accuracy deficit on matched tasks.

\subsection{External and Multi-Signal Validation}
\label{app:external-validation}

This section reports in full the two corroborating checks referenced in Stage 1 of Section~\ref{sec:qa}: a second independent QE metric, and a back-translation round trip compared against a professional human-reference baseline. Both are computed on the same 6,092-pair stratified sample used throughout the corpus's automated screening, excluding Hindi Romanized where noted.

\noindent \textbf{Second-metric convergence.} Table~\ref{tab:comet_comparison} compares the primary screen (\textbf{Unbabel/wmt22-cometkiwi-da}) against a second, independently-trained reference-free metric, \textbf{Unbabel/wmt23-cometkiwi-da-xl}. The two metrics rank translations consistently (Kendall's $\tau$-b $0.46$--$0.57$).

\begin{table}[h]
\centering
\small
\setlength{\tabcolsep}{3pt}
\begin{tabular}{lrrr}
\toprule
\textbf{Lang.} & \textbf{Kiwi $\mu$} & \textbf{2nd $\mu$} & \textbf{$\tau$-b} \\
\midrule
Bengali & 0.854 & 0.678 & 0.479 \\
Hindi   & 0.823 & 0.544 & 0.475 \\
Marathi & 0.762 & 0.613 & 0.574 \\
Tamil   & 0.846 & 0.658 & 0.464 \\
Telugu  & 0.815 & 0.557 & 0.544 \\
\bottomrule
\end{tabular}
\caption{Primary QE metric (Kiwi = wmt22-cometkiwi-da) vs.\ a second, independently-trained metric (2nd = wmt23-cometkiwi-da-xl); $N=1{,}032$ per language.}
\label{tab:comet_comparison}
\end{table}

\noindent \textbf{Back-translation against a human-reference baseline.} We back-translate every Indic translation to English with IndicTrans2 \citep{indictrans2} (a model from a different family than the translator) and score the round trip against the original English with chrF++ and reference-based COMET-22 \citep{comet22}. Table~\ref{tab:backtranslation} reports these alongside the identical pipeline run over IN22-Gen's professional human reference translations \citep{indictrans2}. Hindi Romanized is excluded: IndicTrans2 has no romanized-Hindi source tag.

Our translations round-trip \emph{above} the human references in four of five languages (Bengali $+0.047$ and Hindi $+0.020$ COMET-22, both with bootstrap 95\% CIs excluding zero; Tamil $+0.004$ and Marathi $+0.003$, neither significant), and below only for Telugu ($-0.005$). We therefore treat this as a floor, not a parity or superiority claim. Round-trip score is dominated by how predictable the text is rather than by forward-translation quality: at matched sentence length (27.8 vs.\ 24.8 English words) our questions draw on 1,490 word types across 143k tokens, against IN22-Gen's 9,158 types across 127k (type--token ratio $0.010$ vs.\ $0.072$), because they are synthesized from a small set of SQL skeletons over a fixed schema vocabulary. IN22-Gen's free-domain prose admits several valid renderings per sentence, so a human translator choosing a different-but-correct wording is penalized on the round trip. Two observations confirm the score does not track translation quality on this corpus: Telugu, the one language below the baseline, is also the highest-fertility variant with the largest Indic Gap (Appendix~\ref{app:linguistic-analysis}); and rule-flagged rows score \emph{higher} on chrF++ than clean rows in four of five languages (e.g.\ Hindi $80.2$ vs.\ $75.8$), since retained English prompt-leakage text round-trips perfectly. The check thus bounds gross semantic breakage rather than ranking our translations against professional human work.

\begin{table}[h]
\centering
\small
\setlength{\tabcolsep}{3pt}
\begin{tabular}{lrrrr}
\toprule
\textbf{Lang.} & \textbf{chrF++} & \textbf{Human} & \textbf{COMET-22} & \textbf{Human} \\
\midrule
Bengali & 68.3 & 60.1 & 0.878 & 0.831 \\
Hindi   & 76.8 & 64.4 & 0.905 & 0.885 \\
Marathi & 68.4 & 64.4 & 0.879 & 0.876 \\
Tamil   & 67.3 & 59.3 & 0.873 & 0.870 \\
Telugu  & 71.0 & 64.7 & 0.881 & 0.886 \\
\bottomrule
\end{tabular}
\caption{Back-translation round-trip scores for our translations vs.\ IN22-Gen professional human references under the identical pipeline ($N=1{,}032$ per language for ours; $N=1{,}024$ for IN22-Gen). Our translations score at or above the human references on both metrics in every language except Telugu (COMET-22 $-0.005$). This reflects the low lexical diversity of IndicDBBench's templated questions round-tripping more predictably than IN22-Gen's free-domain prose, not higher translation quality; see text.}
\label{tab:backtranslation}
\end{table}

\noindent \textbf{Complementary defect detection.} Scored against the 1,380 rule-derived hard failures used in Stage 1 screening, the neural QE metrics detect prompt leakage and untranslated (identical-to-English) rows well (AUC $0.59$--$0.89$), since both leave the source string largely or entirely unchanged and are directly visible to a reference-free quality model; back-translation does not (AUC $0.39$--$0.51$, i.e.\ near chance), because prompt leakage retains the correctly-translated question text and back-translates back to fluent, meaning-preserving English regardless. Conversely, the small number of wrong-language-or-script rows in the corpus (translations rendered in a different Indic script than the target language) score normally under CometKiwi ($0.75$--$0.88$, since the QE model is not itself checking the output is in the intended language) but drop to the bottom of the back-translation distribution once decoded with the correct language's source tag, since the model then produces garbage from mismatched-language input. The two signal families are near-independent (Kendall's $\tau$-b $0.02$--$0.14$ between back-translation chrF++ and either QE metric) precisely because they are blind to different defect classes by construction.

\subsection{Evidence Generation and Examples}
\label{app:evidence-generation}

\begin{table*}[t]
\centering
\small
\begin{tabular}{p{0.22\textwidth} p{0.72\textwidth}}
\toprule
\multicolumn{2}{c}{\textbf{Question--Evidence pairs}} \\
\midrule
\textbf{Question} & Provide the area, production, and yield statistics for maize and barley in Chhattisgarh for the year 1970. \\
\textbf{Evidence} & Select maize and barley area, production, yield from fact\_cereals\_minor where dim\_geography.state\_name = `Chhattisgarh` and dim\_year.year = 1970. \\
\midrule
\textbf{Question} & List the station code and the type of water body for all stations located in the state of Assam. \\
\textbf{Evidence} & Assam is a value in dim\_state.state; join dim\_station with dim\_state on state\_id; select station\_code and type\_of\_water\_body. \\
\bottomrule
\end{tabular}
\caption{Question--Evidence pairs for Text-to-SQL reasoning.}
\label{tab:qa_evidence}
\end{table*}

\begin{table*}[t]
\centering
\small
\begin{tabular*}{\textwidth}{@{\extracolsep{\fill}}lccccccc}
\toprule
\textbf{Evidence language} & \textbf{English} & \textbf{Hindi} & \textbf{Bengali} & \textbf{Marathi} & \textbf{Tamil} & \textbf{Telugu} & \textbf{Hinglish} \\
\midrule
Target-language evidence & 25.6 & 27.5 & 24.9 & 24.0 & 25.5 & 24.6 & 24.4 \\
English evidence & 25.6 & 24.0 & 25.6 & 28.0 & 29.0 & 25.7 & 22.9 \\
\bottomrule
\end{tabular*}
\caption{Evidence-language ablation: average execution-accuracy gain (percentage points) from evidence augmentation when the evidence files are generated in the target language versus in English (DeepSeek V3.2). For English questions the two settings coincide; the English-evidence row corresponds to the $\Delta$ column of Table~\ref{tab:evidence_ablation}.}
\label{tab:evidence_language}
\end{table*}

Evidence files are generated automatically with the SEED framework \citep{seed}. For each task, the generator receives the natural-language question, the database schema, and sampled column values from PostgreSQL, and is prompted few-shot with exemplars from BIRD \citep{bird} to produce a short evidence string naming the relevant tables, columns, join keys, and literal values required to answer the question. The gold SQL is never shown to the generator: evidence is produced solely from the question, the schema, and values the generator itself retrieves by probing the database, so it cannot leak the reference query. Evidence is generated separately for every language variant, so each variant's evidence is conditioned on its own question text. The resulting files are supplied identically to all evaluated models in the with-evidence settings and are released with the benchmark. Table~\ref{tab:qa_evidence} shows two representative question--evidence pairs; Appendix~\ref{sec:ablation_statistics} quantifies which components of the generated evidence drive the observed gains.

Because the generator must map phrases of the target-language question onto English schema elements and sketch the tables, joins, and filters the answer requires, the evidence functions as an intermediate representation in which schema grounding and partial query planning are resolved before SQL generation. Although this intermediate text is expressed in English by default, its benefit does not depend on English being the medium: we re-generated the evidence files \emph{in each target language} and repeated the evidence ablation. Table~\ref{tab:evidence_language} compares the resulting gains. The two settings give gains of similar size (22.9--29.0 points; target-language evidence is stronger for Hindi and Hinglish, English evidence for Bengali, Marathi, Tamil, and Telugu, by at most 4.0 points), indicating that the gains stem from the grounding and planning content of the evidence rather than from restoring an English-language bridge to the schema.

\subsection{Generated Evidence Statistics}
\label{sec:ablation_statistics}
\begin{figure*}[t]
\centering
\begin{minipage}[c]{0.46\textwidth}
\centering
\includegraphics[width=\linewidth]{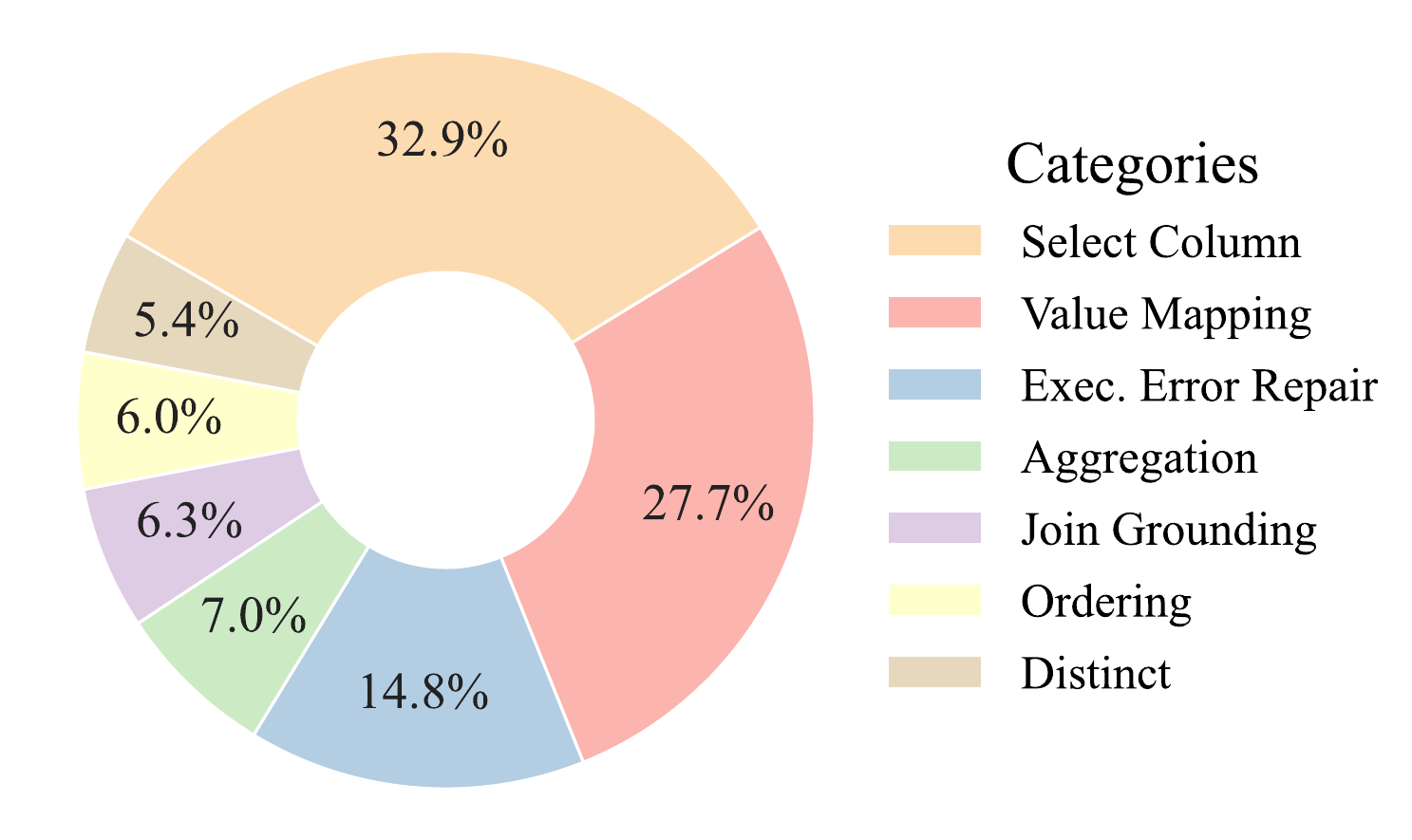}
\end{minipage}\hfill
\begin{minipage}[c]{0.52\textwidth}
\centering
\includegraphics[width=\linewidth]{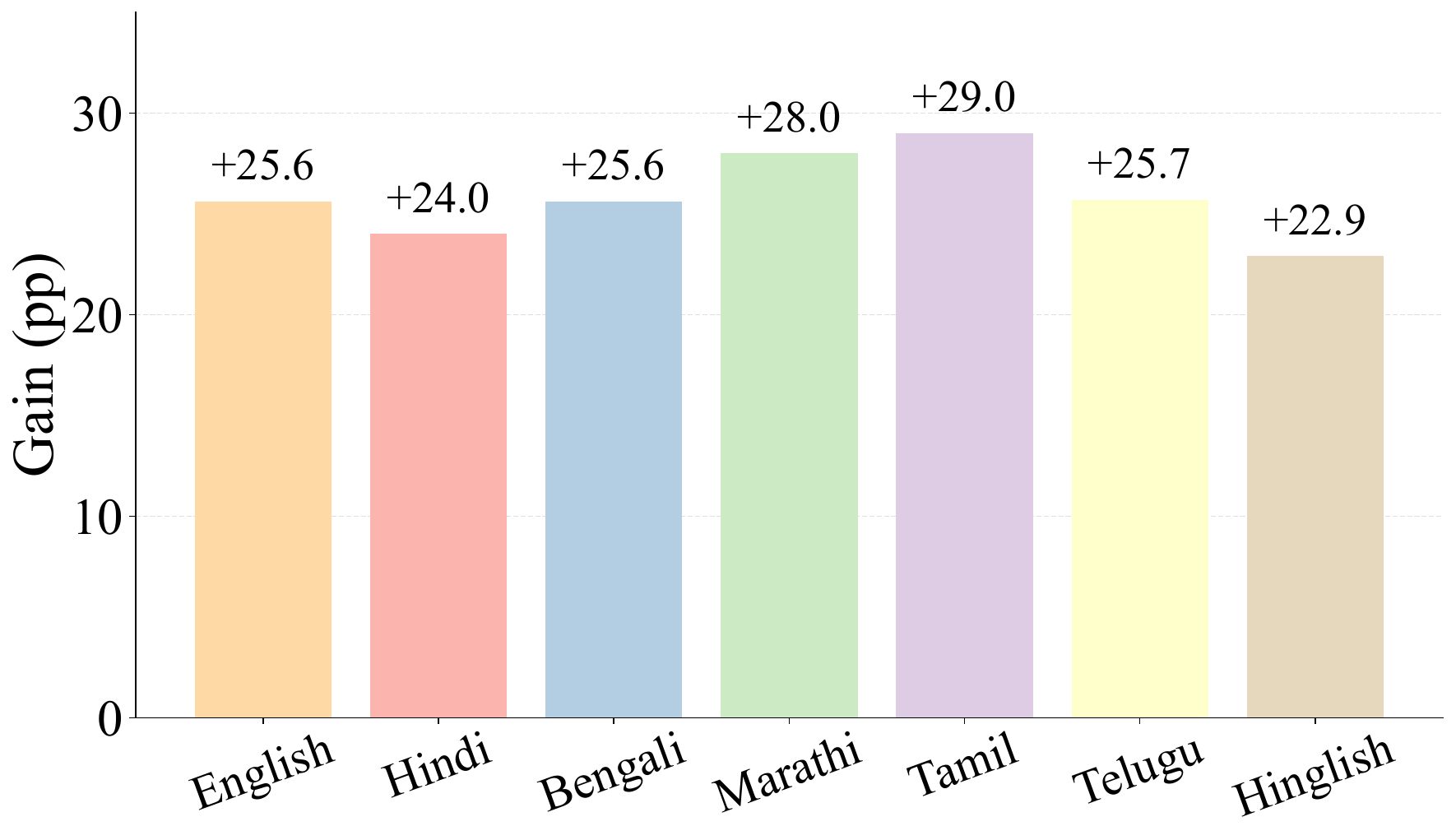}
\end{minipage}
\caption{Impact of evidence files: (Left) distribution of improvement categories among tasks that become correct once evidence is added, (Right) execution-accuracy gains from evidence across languages (DeepSeek V3.2, DIN-SQL; Table~\ref{tab:evidence_ablation}).}
\label{fig:improvements}
\end{figure*}

To characterize \emph{where} evidence helps, we categorize every task that flips from incorrect to correct once evidence is added (DeepSeek V3.2). The left panel shows that grounding-related corrections dominate: correct column selection (32.9\%) and literal value mapping (27.7\%) together account for over 60\% of all improvements, with execution-error repair contributing a further 14.8\%. Compositional categories, namely aggregation (7.0\%), join grounding (6.3\%), ordering (6.0\%), and \texttt{DISTINCT} handling (5.4\%), make up the remainder. This breakdown substantiates the interpretation in Section~\ref{sec:ablation}: evidence acts primarily as a schema- and value-grounding signal rather than as an aid to SQL composition.

The right panel reports the corresponding execution-accuracy gains per language under DIN-SQL (the values underlying Table~\ref{tab:evidence_ablation}). Every language gains between +22.9 and +29.0 points. Relative to English (+25.6), Tamil (+29.0) and Marathi (+28.0) gain most, Telugu (+25.7) and Bengali (+25.6) about the same, and Hindi (+24.0) and Hinglish (+22.9) less. Evidence therefore narrows the gap by at most 3.4 points (Tamil) and widens it for Hindi and Hinglish, so for DeepSeek V3.2 it leaves most of the Indic Gap in place (Section~\ref{sec:ablation}).

\subsection{Oracle Decomposition of Schema Linking}
\label{app:oracle-decomp}

Section~\ref{sec:ablation_statistics} shows descriptively that evidence-driven corrections split between column selection (32.9\%) and literal value mapping (27.7\%). This appendix isolates each sub-component causally, via a controlled oracle ablation.

\noindent\textbf{Conditions.} Four prompts differ only in the oracle information added on top of the (Indic-language) question; question text, instructions, and decoding (temperature 0) are held fixed. \textit{Baseline}: full DDL, no sample rows, no evidence file. \textit{Oracle tables (B)}: DDL pruned to the gold tables the reference query touches, with columns left unpruned. \textit{Oracle entities (C)}: full DDL plus one line listing the gold literal filter values, sorted alphabetically so clause order is not leaked. \textit{Both (B+C)}: pruned schema and value line together. Execution accuracy is scored against gold result sets on a live PostgreSQL instance, as in the main experiments. Unlike SEED evidence files (Appendix~\ref{app:evidence-generation}), these signals intentionally inject gold information: they are diagnostic upper-bound probes, not a deployable prompting strategy.

\noindent\textbf{Sample.} We reuse the 500-task stratified sample of the non-Indic control study (Appendix~\ref{sec:control_appendix}), whose eligible pool requires at least one string literal per gold query (mean 1.5) so that condition C is well-defined on every task. Each task is aligned across all seven language variants and scored under all four conditions on Qwen3~8B and Llama~3.3~70B. The baseline here is deliberately more austere than the main zero-shot condition, so absolute values sit below Table~\ref{tab:full_results} (Qwen3 English 41.8\% vs.\ 52.2\%) and only within-study contrasts are meaningful; the per-language English--Indic gap \emph{ordering} nonetheless tracks the full benchmark (Spearman $\rho = 0.943$).

\begin{table*}[t]
\centering
\small
\setlength{\tabcolsep}{4pt}
\begin{tabular*}{\textwidth}{@{\extracolsep{\fill}}llccccccc}
\toprule
\textbf{Model} & \textbf{Condition} & \textbf{English} & \textbf{Hindi} & \textbf{Bengali} & \textbf{Marathi} & \textbf{Tamil} & \textbf{Telugu} & \textbf{Hinglish} \\
\midrule
\multirow{4}{*}{Qwen3 8B}
 & Baseline                  & 41.8\% & 33.8\% & 30.4\% & 28.6\% & 27.4\% & 27.2\% & 38.9\% \\
 & Oracle tables (B)         & 42.0\% & 36.0\% & 32.0\% & 33.0\% & 29.0\% & 26.8\% & 39.3\% \\
 & Oracle entities (C)       & 45.8\% & 39.0\% & 36.4\% & 35.0\% & 33.8\% & 34.6\% & 44.8\% \\
 & Both (B+C)                & 47.8\% & 40.4\% & 38.4\% & 36.4\% & 36.4\% & 34.6\% & 43.1\% \\
\midrule
\multirow{4}{*}{Llama 3.3 70B}
 & Baseline                  & 47.6\% & 40.0\% & 37.4\% & 33.8\% & 34.4\% & 37.2\% & 44.7\% \\
 & Oracle tables (B)         & 53.6\% & 42.2\% & 44.4\% & 37.2\% & 40.0\% & 42.0\% & 47.7\% \\
 & Oracle entities (C)       & 51.4\% & 47.2\% & 42.0\% & 39.6\% & 39.8\% & 41.2\% & 48.2\% \\
 & Both (B+C)                & 58.2\% & 50.0\% & 46.6\% & 47.4\% & 44.0\% & 45.8\% & 52.9\% \\
\bottomrule
\end{tabular*}
\caption{Execution accuracy (EX) under the oracle-ablation study (500-task stratified sample), by model and condition.}
\label{tab:oracle-decomp-full}
\end{table*}

\begin{table}[t]
\centering
\small
\setlength{\tabcolsep}{3pt}
\resizebox{\columnwidth}{!}{%
\begin{tabular}{lcccc}
\toprule
\textbf{Model} & \textbf{Baseline} & \textbf{Tables (B)} & \textbf{Entities (C)} & \textbf{Both} \\
\midrule
Qwen3 8B      & 12.3 & 10.6 & 10.0 & 10.6 \\
              &      & {\scriptsize [$-6.0$, $+2.8$]} & {\scriptsize [$-6.3$, $+1.7$]} & {\scriptsize [$-6.2$, $+2.7$]} \\
Llama 3.3 70B & 11.0 & 12.4 & 9.4  & 11.4 \\
              &      & {\scriptsize [$-2.0$, $+5.1$]} & {\scriptsize [$-4.4$, $+1.2$]} & {\scriptsize [$-3.4$, $+4.2$]} \\
\bottomrule
\end{tabular}}
\caption{English--Indic gap (English EX minus the mean over the five Indic languages, points) under each oracle condition. Brackets give 95\% task-level bootstrap intervals for the change relative to the baseline gap; every interval includes zero.}
\label{tab:oracle-gap}
\end{table}

\noindent\textbf{Results.} Both oracle signals together raise accuracy in every language, English included, and each alone does so everywhere except Qwen3 Telugu under oracle tables ($27.2\% \rightarrow 26.8\%$; Table~\ref{tab:oracle-decomp-full}), so we evaluate them on the gap itself (Table~\ref{tab:oracle-gap}). For Qwen3~8B the baseline gap of 12.3 points becomes 10.6 with oracle tables, 10.0 with oracle entities, and 10.6 with both: oracle entities lift mean Indic EX by 6.3 points but English by 4.0. For Llama~3.3~70B the gap moves from 11.0 to 12.4, 9.4, and 11.4. No change is significant (2,000 task-level bootstrap resamples), and the largest point estimate, for oracle entities, removes under a fifth of the gap. Per-item McNemar tests (baseline vs.\ oracle) show the signals help individual languages---for Qwen3, oracle entities is significant for all six Indic variants ($p\le0.0073$) and marginal for English ($p=0.065$); for Llama, oracle entities is significant for English and every Indic variant ($p\le0.0365$) and oracle tables for English, Bengali, Tamil, and Telugu ($p\le0.0184$)---but a difference between a significant and a marginal per-language result is not itself evidence of an Indic-specific effect, which is why we test the gap directly. Perfect table and literal-value grounding therefore leaves a large estimated gap for both models. The intervals are wide, however: for Qwen3, oracle entities are compatible with closing about half of the gap ($[-6.3, +1.7]$ points against 12.3), so the study does not rule out a substantial contribution from these signals, and it does not test column or join selection.

\noindent\textbf{Limitations.} Oracle entities is an upper bound on value grounding: supplying `\texttt{'Assam'}' also signals that a filter on that value exists, so condition C is a ceiling for what a transliteration-aware value linker could achieve. The oracle does not cover column selection or join keys, so it bounds table- and value-level grounding only. Only two models are covered.

\subsection{Non-Indic Multilingual Control Study}
\label{sec:control_appendix}

\begin{table*}[t]
\centering
\small
\setlength{\tabcolsep}{4pt}
\begin{tabular*}{\textwidth}{@{\extracolsep{\fill}}llccccccc}
\toprule
 & & & \multicolumn{3}{c}{\textbf{Latin script}} & \multicolumn{3}{c}{\textbf{Non-Latin script}} \\
\cmidrule(lr){4-6}\cmidrule(lr){7-9}
\textbf{Condition} & \textbf{Model} & \textbf{En} & \textbf{Es} & \textbf{Fr} & \textbf{It} & \textbf{Ar} & \textbf{Zh} & \textbf{Ko} \\
\midrule
\multirow{5}{*}{DIN-SQL (w/ ev.)}
 & DeepSeek V3.2 & 66.2 & 70.9 & 67.1 & 69.4 & 67.9 & 64.3 & 65.5 \\
 & Llama 3.3 70B & 66.2 & 71.2 & 70.1 & 71.3 & 69.0 & 70.0 & 69.9 \\
 & MiniMax M2.7  & 69.5 & 71.3 & 67.8 & 69.3 & 67.6 & 66.3 & 66.1 \\
 & Qwen3 8B      & 68.0 & 67.7 & 69.2 & 67.9 & 66.7 & 68.5 & 68.5 \\
 & \emph{Average} & \emph{67.5} & \emph{70.2} & \emph{68.5} & \emph{69.5} & \emph{67.8} & \emph{67.3} & \emph{67.5} \\
\midrule
\multirow{5}{*}{Zero-shot (w/o ev.)}
 & DeepSeek V3.2 & 67.7 & 68.8 & 71.7 & 70.0 & 60.3 & 53.8 & 61.4 \\
 & Llama 3.3 70B & 56.1 & 53.6 & 52.8 & 55.3 & 44.5 & 46.4 & 44.4 \\
 & MiniMax M2.7  & 60.7 & 55.5 & 59.3 & 59.5 & 50.3 & 43.4 & 53.5 \\
 & Qwen3 8B      & 62.3 & 46.9 & 54.5 & 51.0 & 45.7 & 39.7 & 50.0 \\
 & \emph{Average} & \emph{61.7} & \emph{56.2} & \emph{59.6} & \emph{58.9} & \emph{50.2} & \emph{45.8} & \emph{52.3} \\
\midrule
\multirow{5}{*}{Zero-shot (w/ ev.)}
 & DeepSeek V3.2 & 74.3 & 67.1 & 66.4 & 66.1 & 63.7 & 60.2 & 66.9 \\
 & Llama 3.3 70B & 61.3 & 69.6 & 68.8 & 67.5 & 62.6 & 63.1 & 62.8 \\
 & MiniMax M2.7  & 56.1 & 60.0 & 60.0 & 60.4 & 59.7 & 54.7 & 59.5 \\
 & Qwen3 8B      & 63.1 & 67.3 & 66.2 & 64.7 & 64.2 & 54.7 & 65.0 \\
 & \emph{Average} & \emph{63.7} & \emph{66.0} & \emph{65.4} & \emph{64.7} & \emph{62.6} & \emph{58.2} & \emph{63.6} \\
\bottomrule
\end{tabular*}
\caption{Execution accuracy (EX, \%) for the non-Indic control languages under all three prompting conditions, on the 500-task stratified sample. Latin-script controls: Spanish (Es), French (Fr), Italian (It); non-Latin-script controls: Arabic (Ar), Chinese (Zh), Korean (Ko). English (En) is the reference condition.}
\label{tab:control_full}
\end{table*}

This appendix gives per-model, per-language results for the control study summarised in Section~\ref{sec:control_study}, which tests whether the Indic Gap is specific to Indic languages or reflects broader non-English or script effects.

\noindent \textbf{Evaluation sample.}
This is the same 500-task stratified sample used by both the control study (Section~\ref{sec:control_study}) and the oracle-decomposition study (Appendix~\ref{app:oracle-decomp}): 500 instances drawn with a fixed seed (42) from an eligible pool of 725 across 11 databases, stratified jointly by database and difficulty, giving 137 easy, 194 medium, and 169 hard tasks. The eligible pool is restricted to tasks whose gold SQL contains at least one string literal (mean 1.5 per task), a constraint that the oracle-decomposition study needs to make its entity-oracle condition well-defined on every task, and one the control study inherits without loss since it does not condition on difficulty by literal count. Every language variant (Indic, control, and English) is generated from the same 500 selected instances, so all conditions share identical intents, gold SQL, schemas, and difficulty distribution.

\noindent \textbf{Control languages.}
Latin-script non-Indic controls are Spanish, French, and Italian; non-Latin-script non-Indic controls are Arabic, Chinese, and Korean. English is the reference condition. Translations use the same pipeline and prompt as the Indic variants (Appendix~\ref{fig:translation-prompts}), preserving schema tokens and SQL identifiers in English while translating database literal values into the target language. The prompt templates, English schema, and execution-accuracy evaluator are identical to those used for the main results in each corresponding condition.

\noindent \textbf{Group averages.}
Averaging the per-language means in Table~\ref{tab:control_full} gives the Latin and non-Latin group values reported in Table~\ref{tab:control_groups}. Under DIN-SQL with evidence the Latin group reaches $69.4$ and the non-Latin group $67.5$, against $67.5$ for English. Under zero-shot with evidence the corresponding values are $65.4$ and $61.5$ against $63.7$. Under zero-shot without evidence they are $58.2$ and $49.4$ against $61.7$, the only condition in which both control groups fall below English.

\noindent \textbf{Interpretation.}
Two observations follow. First, with evidence the Latin-script controls match or exceed English in both conditions, while the non-Latin controls are level under DIN-SQL and $2.2$ points below English zero-shot (Arabic, Chinese, and Korean each score below English there), so translating the question out of English costs these languages little once grounding is provided. Second, the non-Latin penalty is largest in the zero-shot no-evidence condition, where it reaches $12.3$ points; this is consistent with script and tokenization mismatch mattering most when the model must resolve schema references without explicit grounding signals. With evidence, the controls stay markedly closer to English than the Indic variants, whose average gaps in Table~\ref{tab:full_results} are $3.7$ (DIN-SQL) and $10.4$ points (zero-shot); without evidence the non-Latin gap approaches the Indic one ($12.3$ vs.\ $13.0$), although on a different task sample. The evidence-augmented conditions are therefore the basis for attributing the Indic Gap to factors beyond non-English input and script alone.

\noindent \textbf{Limitations of the control study.}
The control conditions establish a matched comparison across languages, not a causal decomposition of the Indic Gap: language groups differ simultaneously in script, pretraining representation, morphology, and translation quality, and this design cannot separate those factors from one another. The six control languages are all comparatively high-resource, so the comparison bounds the cost of non-English input for well-represented languages rather than for low-resource languages generally.

\subsection{Extended Model Coverage}
\label{app:extra-models}

\begin{table*}[t]
\centering
\small
\setlength{\tabcolsep}{4pt}
\begin{tabular*}{\textwidth}{@{\extracolsep{\fill}}lcccccccc}
\toprule
\textbf{Model} & \textbf{English} & \textbf{Hindi} & \textbf{Bengali} & \textbf{Marathi} & \textbf{Tamil} & \textbf{Telugu} & \textbf{Hinglish} & \textbf{Indic-5 gap} \\
\midrule
Sarvam 105B    & 20.0 & 15.0 & 13.0 & 11.0 & 13.8 & 12.0 & 15.8 &  7.0 \\
Gemma 3 27B    & 49.2 & 32.0 & 30.6 & 26.6 & 29.2 & 34.6 & 43.6 & \textbf{18.6} \\
GPT-5.6 Luna   & 50.8 & 45.4 & 46.0 & 42.8 & 44.6 & 42.6 & 46.0 &  6.5 \\
GPT-5.4 Mini   & 54.2 & 49.0 & 48.4 & 48.6 & 49.8 & 44.8 & 47.5 &  6.1 \\
\bottomrule
\end{tabular*}
\caption{Execution accuracy (EX, \%) for four additional models on the 500-task stratified sample: Sarvam 105B, Gemma 3 27B, GPT-5.6 Luna, and GPT-5.4 Mini.}
\label{tab:extra-models}
\end{table*}

Section~\ref{sec:setup} benchmarks four models chosen on a coverage-first principle. To widen that set we additionally evaluate Sarvam~105B, an open model trained for Indian languages; Gemma~3~27B, drawn from the Indic LLM leaderboard; and two proprietary models, GPT-5.6~Luna and GPT-5.4~Mini (Table~\ref{tab:extra-models}). These runs use a 500-task stratified sample rather than the full benchmark, so their absolute accuracies are not comparable with Table~\ref{tab:full_results}; only the within-model contrast across languages is.

\noindent\textbf{The gap holds across all four.} Defining the Indic-5 gap as English EX minus the mean over Hindi, Bengali, Marathi, Tamil, and Telugu, it is positive for every model: $7.0$ points for Sarvam~105B, $18.6$ for Gemma~3~27B, $6.5$ for GPT-5.6~Luna, and $6.1$ for GPT-5.4~Mini, a mean of $9.6$, the same as the Indic-5 average of the four main models on the full benchmark (Table~\ref{tab:performance-delta}). Hinglish again drops less than the Indic-5 mean under three of the four models, the exception being GPT-5.4~Mini, where the two are within $0.6$ points, consistent with the main results. Neither training targeted at Indian languages (Sarvam~105B) nor proprietary scale (GPT-5.6~Luna, GPT-5.4~Mini) removes the gap, and the largest gap in the table belongs to Gemma~3~27B.

\subsection{Stratified Analyses of the Indic Gap}
\label{app:difficulty-gap}

\begin{table*}[t]
\centering
\small
\begin{tabular*}{\textwidth}{@{\extracolsep{\fill}}lcccccc}
\toprule
 & \multicolumn{3}{c}{\textbf{Zero-shot (w/ evid.)}} & \multicolumn{3}{c}{\textbf{DIN-SQL (w/ evid.)}} \\
\cmidrule(lr){2-4}\cmidrule(lr){5-7}
\textbf{Language} & Easy & Med. & Hard & Easy & Med. & Hard \\
\midrule
English (EX) & 60.6 & 61.5 & 41.7 & 59.7 & 65.9 & 55.8 \\
\midrule
Hindi    & 8.1  & 9.9  & 5.7  & 4.0 & 1.4 & 3.0 \\
Bengali  & 7.4  & 9.9  & 9.3  & 3.6 & 4.7 & 3.5 \\
Marathi  & 8.3  & 11.4 & 10.2 & 7.7 & 7.7 & 6.7 \\
Tamil    & 11.9 & 12.0 & 11.3 & 5.4 & 5.1 & 1.5 \\
Telugu   & 9.2  & 12.8 & 13.0 & 1.4 & 4.7 & 6.0 \\
Hinglish & 10.7 & 7.8  & 6.9  & 1.3 & 2.4 & $-$0.1 \\
\midrule
Average  & 9.3  & 10.6 & 9.4  & 3.9 & 4.3 & 3.4 \\
Rel.\ to English & 15.3\% & 17.3\% & 22.5\% & 6.5\% & 6.6\% & 6.1\% \\
\bottomrule
\end{tabular*}
\caption{Indic Gap (EX drop vs.\ English, percentage points) stratified by task difficulty. Zero-shot: four models, task-matched across languages. DIN-SQL: DeepSeek V3.2 and Qwen3 8B, task-matched, per-task results verified against the aggregate tables. Under zero-shot the relative gap widens with difficulty; DIN-SQL's decomposition both shrinks and flattens it, with Telugu the sole exception.}
\label{tab:difficulty-gap}
\end{table*}

Table~\ref{tab:difficulty-gap} stratifies the Indic Gap by task difficulty under both prompting strategies. Difficulty labels are assigned at synthesis time by the difficulty-calibrated skeletons of the generation pipeline (Section~\ref{sec:tasks}), so every language variant of a task inherits the same tier. Each language cell reports the drop in execution accuracy relative to English \emph{on the identical set of tasks}, so differences in task composition cancel out; the top row gives English's absolute EX as the reference level. Zero-shot results pool all four models; DIN-SQL results pool DeepSeek V3.2 and Qwen3 8B, the models whose per-task logs were retained, with each cell verified by reproducing the corresponding aggregate results (103 of 109 cells within $\pm$1.5 points; the remainder excluded). All comparisons are task-matched within each (model, database) pair across languages.

Under zero-shot prompting, the gap widens with difficulty in relative terms, from 15.3\% of the English score on Easy tasks to 22.5\% on Hard tasks; Telugu's relative deficit doubles (15\%$\rightarrow$31\%), and Bengali, Marathi, and Tamil widen similarly, whereas Hindi and Hinglish remain flat. Under DIN-SQL, whose pipeline adds explicit schema linking and clause-wise decomposition, the gap both shrinks to roughly four points and flattens across difficulty tiers (around 6.5\% at every level); Telugu alone continues to widen (+1.4$\rightarrow$+6.0). This pattern refines our error analysis: the difficulty-dependent component of the gap reflects compositional load interacting with cross-lingual grounding and is largely absorbed by structured decomposition, while the flat residual constitutes a schema-linking floor that persists even with evidence and decomposition. Marathi is the notable outlier, retaining the largest residual gap under DIN-SQL (around 7 points at every tier); Appendix~\ref{app:translation-noise} shows this floor is not attributable to translation noise.

\begin{table*}[t]
\centering
\small
\begin{tabular*}{\textwidth}{@{\extracolsep{\fill}}lccc}
\toprule
 & \textbf{Avg.\ gap (pp)} & \textbf{English EX (\%)} & \textbf{Relative gap} \\
\midrule
\multicolumn{4}{l}{\emph{Join width}} \\
\quad 0--1 joins & +9.5 & 69.0 & 13.7\% \\
\quad 2 joins & +12.5 & 53.4 & 23.4\% \\
\quad 3+ joins & +7.2 & 33.4 & 21.7\% \\
\midrule
\multicolumn{4}{l}{\emph{Aggregation complexity}} \\
\quad No aggregation & +7.2 & 65.2 & 11.1\% \\
\quad GROUP BY & +14.0 & 67.6 & 20.7\% \\
\quad GROUP BY + HAVING$^\dagger$ & +12.9 & 52.4 & 24.5\% \\
\midrule
\multicolumn{4}{l}{\emph{Foreign-key hops traversed}} \\
\quad 0 hops & +7.0 & 52.1 & 13.4\% \\
\quad 1 hop & +10.9 & 59.1 & 18.5\% \\
\quad 2 hops & +11.9 & 54.0 & 22.1\% \\
\bottomrule
\end{tabular*}
\caption{Indic Gap by structural complexity of the gold query. Gaps are task-matched execution-accuracy drops vs.\ English on the same tasks, averaged over six language variants and four models (zero-shot with evidence); relative gap = average gap / English EX. Foreign-key hops denote the longest foreign-key path traversed by the gold query; sparsely populated buckets (3+ hops, aggregates without \texttt{GROUP BY}) are omitted. $^\dagger$Includes queries with \texttt{HAVING} or two or more aggregates.}
\label{tab:structural-gap}
\end{table*}

\noindent \textbf{Structural axes.} The gap also grows along every structural axis of the gold query (Table~\ref{tab:structural-gap}); these comparisons are task-matched and pooled over the four models in the zero-shot with-evidence setting. Relative to English, the average Indic deficit roughly doubles from single-join to two-join queries (13.7\% to 23.4\%), from non-aggregating to grouping queries (11.1\% to 20.7--24.5\%), and from zero to two foreign-key hops (13.4\% to 22.1\%). The foreign-key result directly ties the Indic Gap to the administrative hierarchies that motivate the benchmark: the deeper the hierarchy a query must traverse, the larger the cross-lingual penalty. The aggregation result independently corroborates our error taxonomy, in which aggregation and \texttt{GROUP BY} failures form the largest error class.

\subsection{Translation Noise and the Marathi Flag Rate}
\label{app:translation-noise}

\begin{table*}[t]
\centering
\small
\begin{tabular*}{\textwidth}{@{\extracolsep{\fill}}lcccc}
\toprule
 & & & \multicolumn{2}{c}{\textbf{Indic Gap vs English (pp)}} \\
\cmidrule(lr){4-5}
\textbf{Language} & \textbf{Mean COMET} & \textbf{Flagged ($\tau{=}0.737$)} & clean subset & flagged subset \\
\midrule
Bengali & 0.854 & 1.5\% & +10.8 & +7.5 \\
Hindi & 0.823 & 2.8\% & +9.3 & +3.2 \\
Tamil & 0.846 & 3.1\% & +12.9 & +3.9 \\
Telugu & 0.815 & 6.2\% & +13.2 & +14.7 \\
\textbf{Marathi} & \textbf{0.762} & \textbf{26.0\%} & \textbf{+12.0} & \textbf{+13.3} \\
Hinglish$^*$ & 0.669 & 73.2\% & -- & -- \\
\bottomrule
\end{tabular*}
\caption{COMET-QE screening and its relation to the Indic Gap. Gaps are task-matched EX drops vs.\ English on the same tasks, pooled over four models (zero-shot with evidence). Marathi's gap is unchanged on the COMET-screen-passing subset, indicating its elevated flag rate reflects COMET-QE calibration rather than translation damage. $^*$Hinglish COMET scores are unreliable on romanized text and were excluded from thresholding.}
\label{tab:marathi-comet}
\end{table*}

Marathi exhibits a markedly higher COMET-QE flagging rate (26.0\% at $\tau = \mu - 1\sigma$) than the other languages (1.5--6.2\%; Table~\ref{tab:marathi-comet}). Two task-matched analyses indicate this reflects metric calibration rather than translation damage. First, flagged Marathi items show no execution-accuracy deficit relative to clean items on matched tasks (46.2\% vs.\ 44.2\%), which would be expected if the flagged translations were semantically broken. Second, the Marathi--English gap on the COMET-screen-passing subset (COMET $\geq \tau$) is +12.0 points, essentially identical to the flagged subset (+13.3), so Marathi's performance drop cannot be attributed to residual translation noise. This is consistent with prior evidence that the correlation between COMET and human judgments varies across Indian languages \citep{indicmteval}. A post-hoc artifact sweep bounds the influence of residual translation noise on all per-language gaps to within $\pm$1.3 points.

\begin{table}[t]
\centering
\small
\setlength{\tabcolsep}{4pt}
\begin{tabular}{lcccc}
\toprule
\textbf{COMET third} & \textbf{$n$} & \textbf{En EX} & \textbf{Gap} & \textbf{Loss rate} \\
\midrule
Worst   & 5,432 & 58.9 & 11.1 & 25.8 \\
Middle  & 5,446 & 57.4 & 10.7 & 25.9 \\
Best    & 5,450 & 53.1 & 12.2 & 31.6 \\
\midrule
All     & 16,328 & 56.5 & 11.3 & 27.7 \\
\bottomrule
\end{tabular}
\caption{Indic Gap by translation quality. Within each language, questions are split into three equal groups by CometKiwi score (worst, middle, best). The five Indic languages and four models (zero-shot with evidence) are pooled. Gap = English EX minus Indic EX on the same tasks, in points. Loss rate = share of tasks solved in English whose Indic version fails.}
\label{tab:comet-terciles}
\end{table}

\noindent\textbf{Does translation quality explain the gap?} If the Indic Gap came from poor translations, it should be largest for the worst-translated questions and shrink for the best. We test this on every scored translation. For each Indic question, we take its CometKiwi score and the result of the same task in English and in the Indic language. We do this for all four models under zero-shot with evidence (16,328 question pairs over 991 tasks). Each language's questions are split into three equal groups by score; splitting within each language keeps Marathi's lower COMET scores from skewing the groups. Every comparison uses the same task and the same gold SQL, so query difficulty is held fixed.

The gap does not shrink as translations get better (Table~\ref{tab:comet-terciles}): it is 11.1 points for the worst third and 12.2 for the best. Tasks solved in English also fail in the Indic version slightly \emph{more} often when the translation is best (31.6\% vs.\ 25.8\%). The same holds for Hindi, Marathi, Tamil, and Telugu (worst vs.\ best third: 7.6 vs.\ 9.8, 12.9 vs.\ 12.9, 11.6 vs.\ 14.7, 12.0 vs.\ 15.1). Bengali is the exception (11.2 vs.\ 8.4). A regression gives the same answer: it controls for whether the English version was solved and for language and model. A one-standard-deviation increase in COMET changes Indic accuracy by $-0.9$ points (95\% interval $[-2.0, +0.1]$), so better translations bring no measurable gain.

Two caveats apply. COMET is only a proxy for translation quality, and the best-translated questions happen to be slightly harder in English (53.1\% vs.\ 58.9\%). The analysis therefore rules out a \emph{strong} effect of translation quality on the gap, not every effect.

\subsection{Tokenization Fertility and the Indic Gap}
\label{app:linguistic-analysis}

\begin{table*}[t]
\centering
\small
\begin{tabular*}{\textwidth}{@{\extracolsep{\fill}}lcccccc}
\toprule
 & \multicolumn{4}{c}{\textbf{Token fertility ($\times$ English)}} & & \\
\cmidrule(lr){2-5}
\textbf{Language} & Llama 3.3 & Qwen3 & DeepSeek & MiniMax & \textbf{Avg.\ fert.} & \textbf{Avg.\ drop (pp)} \\
\midrule
Hindi    & 2.64 & 4.53 & 2.90 & 2.98 & 3.26 & 9.1 \\
Bengali  & 6.03 & 5.15 & 2.08 & 3.92 & 4.30 & 7.7 \\
Marathi  & 2.85 & 4.67 & 3.05 & 3.31 & 3.47 & 9.9 \\
Tamil    & 7.90 & 6.29 & 3.22 & 4.13 & 5.39 & 8.3 \\
Telugu   & \textbf{8.66} & \textbf{7.21} & \textbf{3.90} & \textbf{4.40} & \textbf{6.04} & \textbf{12.8} \\
Hinglish & 1.34 & 1.33 & 1.36 & 1.36 & 1.35 & 6.3 \\
\bottomrule
\end{tabular*}
\caption{Token fertility by language and tokenizer: tokens required for the same question relative to its English source, on aligned pairs. Avg.\ drop is the per-language mean EX drop vs.\ English across all models and settings (Table~\ref{tab:full_results}).}
\label{tab:fertility}
\end{table*}

\begin{figure}[t]
\centering
\includegraphics[width=\columnwidth]{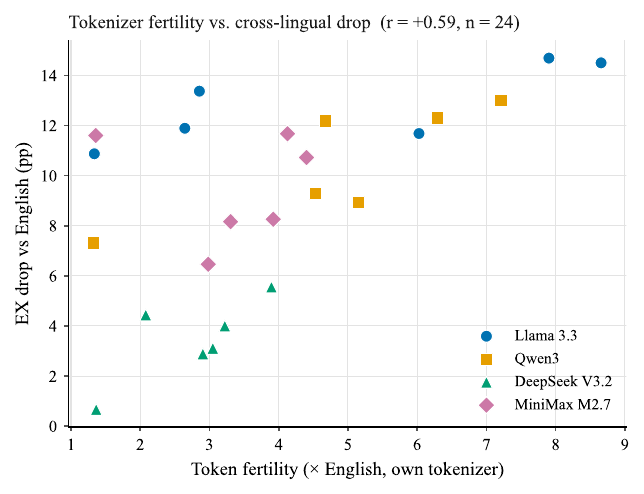}
\caption{Token fertility versus cross-lingual EX drop for all 24 model--language pairs (each model's own tokenizer; task-matched drops, zero-shot with evidence). Pearson $r = +0.59$; within every model, higher-fertility languages show larger drops.}
\label{fig:fertility-scatter}
\end{figure}

To move beyond ``linguistic distance'' as an explanation, we quantify tokenization pressure directly. For every aligned question pair we compute \emph{token fertility}: the number of tokens a model's tokenizer requires for the Indic question relative to its English source (Table~\ref{tab:fertility}). The same question costs on average 6.0$\times$ more tokens in Telugu than in English (8.7$\times$ under Llama's tokenizer), 5.4$\times$ in Tamil, and only 1.35$\times$ in Hinglish. Concretely: the average English question in our corpus occupies 35 tokens under Llama's tokenizer, while the same questions average 281 tokens in Telugu and 259 in Tamil, at 0.66 and 0.79 \emph{characters} per token, meaning most characters are split into multiple byte-level fragments. The content words that must be matched against English schema identifiers therefore reach the model as long runs of sub-lexical pieces rather than as coherent lexical units, which is the mechanism by which fertility taxes schema linking. Fertility correlates with the observed per-language drop both across languages ($r = +0.70$) and across all 24 model--language pairs, using each model's own tokenizer and task-matched drop ($r = +0.59$, Figure~\ref{fig:fertility-scatter}). With six variants the cross-language correlation is not significant on its own ($p \approx 0.12$) and is partly carried by Hinglish ($r = +0.47$ over the five Indic languages), and the 24 model--language pairs are not independent; we therefore read fertility as one contributing factor rather than a sufficient explanation. The model dimension is consistent: DeepSeek V3.2 has the most efficient Indic tokenization (2.1--3.9$\times$ English) and the smallest cross-lingual gaps (+0.7 to +5.6 points), whereas Llama 3.3 has the least efficient (up to 8.7$\times$) and the largest (+10.9 to +14.7 points). Fertility is not the whole story (Llama's Hindi gap is large despite modest fertility, indicating that pretraining exposure contributes independently), but agglutinative morphology visibly compounds the gap, consistent with Telugu being the only language whose deficit continues to widen under DIN-SQL (Appendix~\ref{app:difficulty-gap}). A controlled constituent-order experiment (Appendix~\ref{app:sov}) further rules out SOV word order itself as a driver of the gap.

\subsection{Constituent-Order (SOV) Experiment}
\label{app:sov}

\begin{table*}[t]
\centering
\small
\begin{tabular*}{\textwidth}{@{\extracolsep{\fill}}llccccc}
\toprule
\textbf{Model} & \textbf{Language} & \textbf{n} & \textbf{EX orig.} & \textbf{EX reord.} & \textbf{$\Delta$ (pp)} & \textbf{95\% CI / $p$} \\
\midrule
\multirow{7}{*}{Llama 3.3 70B}
 & English  & 200 & 48.0 & 47.5 & $-$0.5 & [$-$6.0, +5.0] / 1.00 \\
 & Hindi    & 193 & 43.0 & 40.9 & $-$2.1 & [$-$6.7, +2.6] / 0.52 \\
 & Bengali  & 187 & 34.8 & 34.2 & $-$0.5 & [$-$5.9, +4.8] / 1.00 \\
 & Marathi  & 184 & 35.3 & 33.2 & $-$2.2 & [$-$7.6, +3.3] / 0.54 \\
 & Tamil    & 189 & 34.9 & 32.3 & $-$2.6 & [$-$7.9, +2.6] / 0.44 \\
 & Telugu   & 195 & 40.0 & 37.9 & $-$2.1 & [$-$7.7, +3.6] / 0.59 \\
 & Hinglish & 179 & 49.7 & 45.8 & $-$3.9 & [$-$9.5, +1.7] / 0.25 \\
\midrule
\multirow{7}{*}{DeepSeek V3.2}
 & English  & 200 & 60.0 & 56.0 & $-$4.0 & [$-$9.0, +1.0] / 0.19 \\
 & Hindi    & 193 & 58.0 & 52.8 & $-$5.2 & [$-$9.8, $-$0.5] / 0.052 \\
 & Bengali  & 187 & 53.5 & 46.5 & $-$7.0 & [$-$11.2, $-$2.7] / \textbf{0.002} \\
 & Marathi  & 184 & 47.3 & 44.0 & $-$3.3 & [$-$8.2, +1.6] / 0.29 \\
 & Tamil    & 189 & 48.7 & 47.6 & $-$1.1 & [$-$6.3, +4.2] / 0.85 \\
 & Telugu   & 195 & 54.4 & 47.2 & $-$7.2 & [$-$12.3, $-$2.6] / \textbf{0.007} \\
 & Hinglish & 179 & 54.2 & 55.3 & +1.1 & [$-$3.4, +5.6] / 0.80 \\
\midrule
\multirow{7}{*}{Qwen3 8B}
 & English  & 200 & 29.0 & 25.0 & $-$4.0 & [$-$9.0, +1.0] / 0.15 \\
 & Hindi    & 193 & 20.2 & 18.1 & $-$2.1 & [$-$6.7, +2.6] / 0.50 \\
 & Bengali  & 187 & 16.6 & 20.3 & +3.7 & [+0.5, +7.5] / 0.065 \\
 & Marathi  & 184 & 10.3 & 11.4 & +1.1 & [$-$2.2, +4.3] / 0.75 \\
 & Tamil    & 189 & 16.9 & 14.8 & $-$2.1 & [$-$6.9, +2.6] / 0.50 \\
 & Telugu   & 195 & 16.4 & 14.4 & $-$2.1 & [$-$6.2, +1.5] / 0.42 \\
 & Hinglish & 179 & 25.1 & 22.3 & $-$2.8 & [$-$7.3, +1.7] / 0.36 \\
\bottomrule
\end{tabular*}
\caption{Constituent-order experiment: execution accuracy on original vs.\ order-rewritten questions for three model families (zero-shot, no evidence; paired on identical tasks over four databases; Qwen3 evaluated in non-thinking mode). English is rewritten from SVO into condition-first, verb-final form; Indic variants are rewritten with the request phrase fronted. $\Delta$ = reordered $-$ original; CIs are bootstrap 95\% intervals over tasks and $p$ is an exact McNemar test on discordant pairs; significant values in bold.}
\label{tab:sov}
\end{table*}

All five Indic languages, and the Hindi grammatical frame of Hinglish, place constituents in subject--object--verb (SOV) order, whereas English is subject--verb--object: the request and conditions of a question therefore reach a left-to-right decoder in a different sequence. To test whether SOV word order itself contributes to the Indic Gap, we rewrite each question's constituent order in both directions and re-evaluate on identical tasks (Table~\ref{tab:sov}). English questions are rewritten into condition-first, verb-final (SOV-like, topicalized) English; Indic and Hinglish questions are rewritten with the request phrase moved to the front (English-like information order). Rewrites are produced by Gemini 2.5 Flash-Lite under order-only constraints and pass automated integrity checks (numeric literals, entities, and script preserved); 94.8\% of rewrites pass and only passing items are evaluated (previously identified instruction-leak prefixes were removed from both conditions). We evaluate 200 tasks per language over four databases in the zero-shot no-evidence setting, for three model families: Llama 3.3 70B, DeepSeek V3.2, and Qwen3 8B (non-thinking mode).

\noindent\textbf{Example.} An English question and its order rewrite from the released audit file:
\begin{quote}\small
\textbf{Original (SVO):} ``Identify the top 5 rural sub-districts in the Andaman and Nicobar Islands with the highest total allopathic outpatient attendance for the year 2020.''

\textbf{Reordered (verb-final):} ``For the year 2020, in the Andaman and Nicobar Islands, the top 5 rural sub-districts with the highest total allopathic outpatient attendance, identify them.''
\end{quote}
\begin{figure}[t]
\centering
\includegraphics[width=\columnwidth]{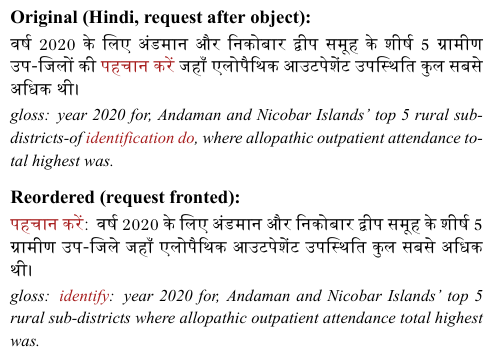}
\caption{The Hindi variant of the same example task before and after intent-fronting. The request phrase (red) moves from its natural post-object position to the front; every content word, number, and entity is unchanged. Italic lines are interlinear glosses preserving the Hindi word order in English.}
\label{fig:sov-hindi}
\end{figure}

\noindent The request (\emph{identify}) moves from first to last position and the conditions (year, region) move to the front, while every entity, number, and modifier is preserved, exactly the information order an SOV language presents. The Indic rewrites apply the mirror-image transformation: Figure~\ref{fig:sov-hindi} shows the Hindi variant of the same task, whose original places the request phrase after its object and whose rewrite moves it to the front, giving the Indic sentence English-like information order with every content word unchanged.

The SOV hypothesis predicts that intent-fronting should improve Indic accuracy. It never does: across the 18 Indic and Hinglish conditions spanning three model families, not one shows a significant gain (the only positive deltas, Qwen3 on Bengali and Marathi and DeepSeek on Hinglish, are not significant), and no Llama or Qwen delta is significant in any language. Where reordering does have a significant effect it \emph{degrades} accuracy: DeepSeek V3.2 loses 7.0 points on Bengali ($p=0.002$) and 7.2 on Telugu ($p=0.007$). Reordered \emph{English}, where language, script, and tokenizer are held constant, drops under all three models ($-$0.5 to $-$4.0, n.s.). This pattern indicates sensitivity to natural phrasing rather than to any particular constituent order; because the same fluency penalty applies to the Indic rewrites, the null on intent-fronting is conservative. We conclude that constituent order is not a meaningful driver of the Indic Gap, reinforcing the grounding- and tokenization-based account (Appendices~\ref{app:difficulty-gap} and~\ref{app:linguistic-analysis}).

\subsection{Generated Schema Example}
\label{app:schema-example}

\begin{figure*}[t]
\centering
\includegraphics[width=0.95\textwidth]{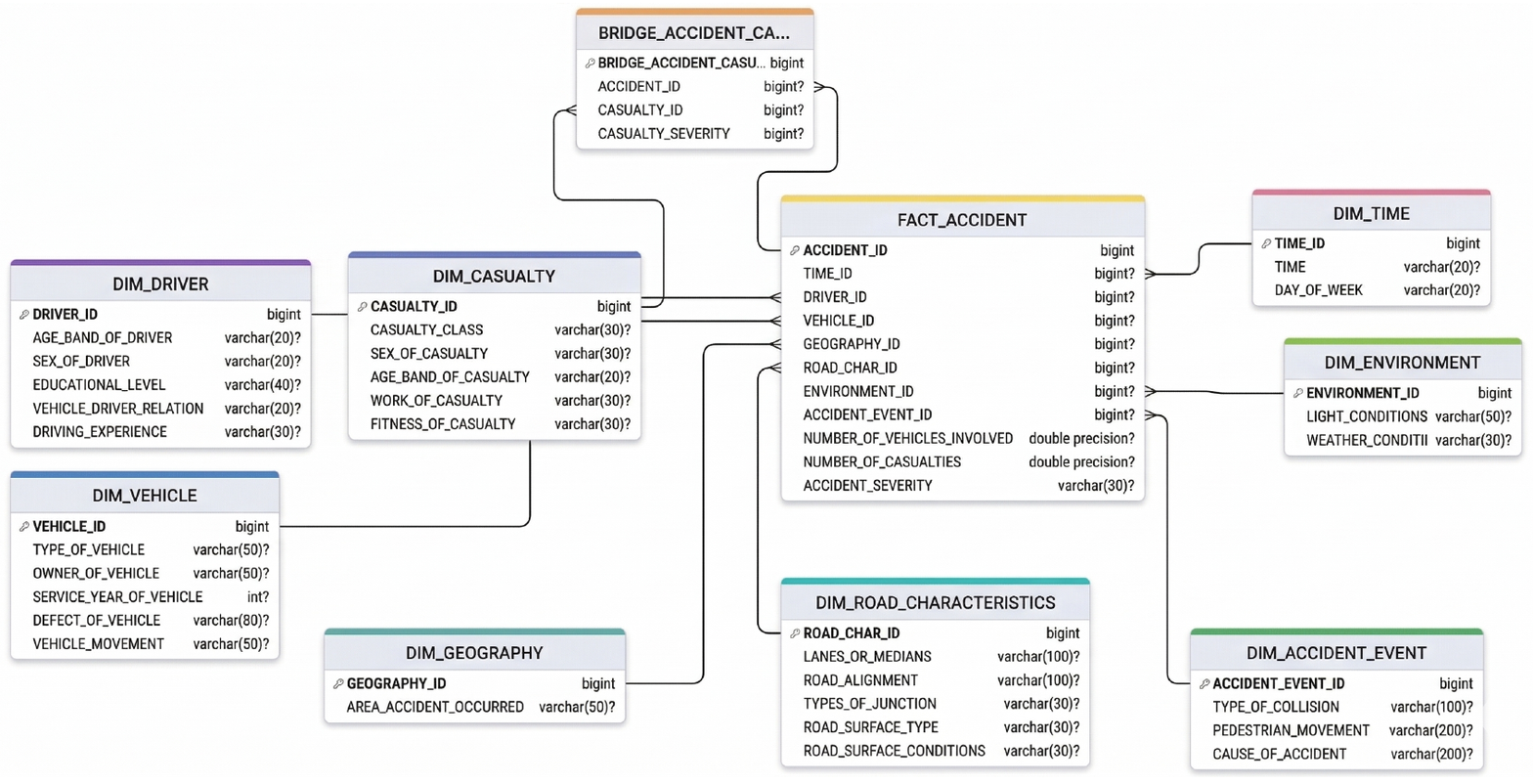}
\caption{Generated schema for the road-accidents database (\texttt{INDIA\_ROAD\_ACCIDENTS\_DATASET\_2001}): a star schema with a central fact table, eight dimension tables, and a bridge table resolving the many-to-many accident--casualty relationship.}
\label{fig:example_schema}
\end{figure*}

Figure~\ref{fig:example_schema} shows a representative output of the Architect--Auditor--Refiner pipeline (Section~\ref{sec:schema}), produced from a single monolithic source CSV. The result is a star schema of \textbf{10 tables, 51 columns, and 9 foreign-key constraints} holding 48{,}051 rows: the central \texttt{FACT\_ACCIDENT} table (20{,}000 accidents, 11 columns) references seven dimension tables (driver, vehicle, geography, time, environment, road characteristics, and accident event), while \texttt{BRIDGE\_ACCIDENT\_CASUALTY} (20{,}000 links) resolves the many-to-many relationship between accidents and casualties.

Two properties make such schemas challenging evaluation targets. First, dimension cardinalities span three orders of magnitude, from 13 rows (\texttt{DIM\_GEOGRAPHY}) to 4{,}419 (\texttt{DIM\_TIME}), so value grounding ranges from trivial to high-cardinality lookup within a single database. Second, because every quantitative attribute lives in the fact table and every descriptive attribute in a dimension, almost any natural question (e.g., ``accidents involving drivers with 1--2 years of experience'' or ``casualty severity by age band'') forces at least one fact--dimension join, and questions touching casualties must additionally traverse the bridge table; the deepest gold queries over this schema combine six join operations.

\subsection{Prompt Template}\label{app:prompt_template}

\Needspace{8\baselineskip}
\subsubsection{Prompt for Schema Linking}

\begin{tcolorbox}[
    title=Schema Linking Prompt,
    colback=white,%
    colframe=gray!75!black,%
    fonttitle=\bfseries,%
    enhanced,%
    breakable%
]
You are a smart and responsible  PostgreSQL expert. Assist in identifying the database tables 
and columns involved in natural language queries.

\vspace{1em}

\#\#\# Instruction:

Your task is to analyze the provided database schema, comprehend the posed question, and leverage
the hint to identify which tables are needed to generate a SQL query for answering the question. The returned JSON format must strictly adhere to the following specifications:

\begin{lstlisting}
{
    "tables": [
        {
            "table": "table name",
            "columns": ["relevant column 1", "relevant column 2", ...]
        },
        ...
    ]
}
\end{lstlisting}

Each relevant column must belong to its respective table, and the output JSON object must be
wrapped in a code block using \texttt{```json```}.
Please note that each table and column comes with detailed description information and example
values for reference.

\vspace{1em}

\#\#\# Database schema:

\{schema\_str\}

\vspace{1em}

\#\#\# User question:

\{query\}

\vspace{1em}

\#\#\# Hint:
\{evidence\}
\end{tcolorbox}
\label{fig:schema_linking}

\Needspace{8\baselineskip}
\subsubsection{Prompt for Basic SQL Generation Pipeline}

\begin{tcolorbox}[
    title=Basic SQL Generation Pipeline,
    colback=white,          
    colframe=gray!75!black, 
    fonttitle=\bfseries,    
    enhanced,               
    breakable               
]
You are a intelligent and responsible PostgreSQL expert. 

\vspace{1em}

\#\#\# Instruction:

You need to read the database schema to generate SQL query for the user question.

The outputted SQL must be surrounded by \verb|```sql```| code block.

\vspace{1em}

\#\#\# Database Schema:

\{schema\}

\vspace{1em}

\#\#\# Hint:

\{evidence\}

\vspace{1em}

\#\#\# User Question:

\{query\}

\vspace{1em}

The outputted SQL must be surrounded by \verb|```sql```| code block.

\end{tcolorbox}

\label{fig:prompt-basic-generator}

\Needspace{8\baselineskip}
\subsubsection{Prompt for Divide-and-Conquer Chain-of-Thought}

\begin{tcolorbox}[
    title=Divide Prompt,
    colback=white,%
    colframe=gray!75!black,%
    fonttitle=\bfseries,%
    enhanced,%
    breakable%
]
You are a smart and responsible PostgreSQL expert. Given a database schema and a question, users 
want to know the corresponding SQL query. Your task is to understand the database schema and 
question, and decompose the question into sub-questions so user can better understand it. 
Each sub-question is enclosed in \textless\textless\textgreater\textgreater. Here is an example 
for reference:

\vspace{1em}

\#\#\# Example:

\vspace{1em}

\#\# Given the database schema:

\{example\_database\_schema\}

\vspace{1em}

\#\# Question: 

\{example\_question\}

\vspace{1em}

\#\# Decompose the Question into sub-questions, each sub-question is enclosed in 
\textless\textless\textgreater\textgreater:

Sub-question 1: \textless\textless \{sub question 1\}\textgreater\textgreater

Sub-question 2: \textless\textless \{sub question 2\}\textgreater\textgreater 

Sub-question 3: \textless\textless \{sub question 3\}\textgreater\textgreater 

\vspace{1em}

\#\#\# Your task: decompose the question into sub-questions.

\vspace{1em}

\#\# Given the database schema:

\{schema\}

\vspace{1em}

\#\# Question: 

\{query\}

\vspace{1em}

\#\# Hint:

\{evidence\}

\vspace{1em}

\#\# Decompose the Question into sub-questions, each sub-question is enclosed in 
\textless\textless\textgreater\textgreater:
\end{tcolorbox}

\label{fig:prompt-divide}

\begin{tcolorbox}[
    title=Conquer Prompt,
    colback=white,        
    colframe=gray!75!black,
    fonttitle=\bfseries,    
    enhanced,%
    breakable%
]
You are a smart and responsible  PostgreSQL expert. Given a database schema and a question, 
your tasks are:

\vspace{1em}

1. Parse user questions: Use natural language processing (NLP) techniques to parse user 
questions and extract query requirements and conditions.

\vspace{1em}

2. Analyze database schema: Based on the database schema, understand the fields and 
relationships of the table, and build the basic framework of the SQL query.

\vspace{1em}

3. Check sample data: Analyze the data characteristics based on the first three rows of 
the table values to help determine how to construct query conditions and filter results.

\vspace{1em}

4. Generate SQL query: Based on user questions, query requirements and conditions, 
database schema, and sample data, build a complete SQL query.

\vspace{1em}

5. Verification and optimization: Check whether the generated SQL query is logical and 
optimize it if necessary.

\vspace{1em}

\#\#\# Database Schema:

\{schema\}

\vspace{1em}

\#\#\# Examples:

\{examples\}

\vspace{1em}

\#\#\# Question:

\{query\}

\vspace{1em}

\#\#\# Hint:

\{evidence\}

\vspace{1em}

Please generate the corresponding SQL query. SQL must be surrounded by \verb|```sql```| code block.
\end{tcolorbox}

\label{fig:prompt-conquer}

\begin{tcolorbox}[
    title=Assemble Prompt,
    colback=white,%
    colframe=gray!75!black,%
    fonttitle=\bfseries,%
    enhanced,%
    breakable%
]
You are a smart and responsible PostgreSQL expert. Given a database schema and a question, 
users want to know the corresponding SQL query.

\vspace{1em}

\#\#\# Instructions:

We have decomposed the main question into sub-questions, now your task is based on the SQL querys 
for corresponding sub-questions, assemble the final SQL for the main question:

1. Understand the database schema and the main question;

2. Read and analyze each sub-question and corresponding SQL query;

3. Analyze the relationship between sub-questions and the main question in order to assemble them properly;

4. Generate the final SQL for the main question and optimize it if needed.

\vspace{1em}

\#\#\# Database Schema:

\{schema\}

\vspace{1em}

\#\#\# Main question:

\{query\}

\vspace{1em}

\#\#\# Hint:

\{evidence\}

\vspace{1em}

\#\#\# Sub-questions and corresponding output, including SQL querys and explanation:

\{subs\}

\vspace{1em}

Based on the SQL querys for corresponding sub-questions, generate the final SQL for the main question 
in the end of your response, SQL must be surrounded by \verb|```sql```| code block.
\end{tcolorbox}

\label{fig:prompt-assemble}

\Needspace{8\baselineskip}
\subsubsection{DIN-SQL Prompt}

\begin{tcolorbox}[
    title=Debugger Prompt,
    colback=white,%
    colframe=gray!75!black,%
    fonttitle=\bfseries,%
    enhanced,%
    breakable%
]
 For the given question, use the provided tables, columns, foreign keys, and primary keys to fix the given PostgreSQL QUERY for any issues. If there are any problems, fix them. If there are no issues, return the PostgreSQL QUERY as is.\\\\
Use the following instructions for fixing the SQL QUERY:\\\\
1) Pay attention to the columns that are used for the JOIN by using the Foreign\_keys.\\\\
2) Use DESC and DISTINCT when needed\\\\
3) Pay attention to the columns that are used for the GROUP BY statement\\\\
4) Pay attention to the columns that are used for the SELECT statement.\\\\
5) Only change the GROUP BY clause when necessary (Avoid redundant columns in GROUP BY).\\\\
6) The question may be in non-english language, the sql query has to be in english.\\\\
7) Don't include back-ticks around table names or columns names in the SQL query
\end{tcolorbox}
\label{fig:prompt-online-synthesis}

\subsection{DSQG-Syn enhanced prompts}
\label{fig:dsqg-syn-prompts}
\begin{tcolorbox}[
    title=DSQG-Syn Skeleton Generation Prompt,
    colback=white,
    colframe=gray!75!black,
    fonttitle=\bfseries,
    enhanced,
    breakable
]
Please generate \{num\_skeletons\} SQL templates based on the given question and schema. Ensure that a mix of SQL clauses are included, such as SELECT, FROM, JOIN, WHERE, GROUP BY, ORDER BY, and HAVING.

\vspace{1em}

\#\#\# Instruction:

1. Use `col\_\#` for column names.\\
2. Use `table\_\#` for table names.\\
3. Use `value\_\#` for constant values.\\
4. Follow the difficulty guidance in \{difficulty\_instructions\}.\\

\vspace{1em}

\#\#\# Prompt Body:

\begin{lstlisting}
Please generate {num_skeletons} SQL templates based on the
given
question and schema. Ensure that a mix of SQL clauses are
included, such as SELECT, FROM, JOIN, WHERE, GROUP BY, ORDER BY,
and HAVING. Use placeholders for specific table and column
names as follows:

1. Use col_# for column names.
2. Use table_# for table names.
3. Use value_# for constant values.

{difficulty_instructions}

Example:
Input:
{"question": "Show me the redshift of spectroscopic object with 
subclass of STARFORMING"}
Schema:

CREATE TABLE specobj (
specobjid number Example Values[(Decimal('299489952322840576'),
), ...],
subclass text Example Values[(None,), ('BROADLINE',),
('STARFORMING',)],
z number Example Values[(7.01124,),
(0.00415325,), (0.00415325,)],
. . . . . .
primary key (specobjid)
)
Output:
{
  "templates": [
    {"template": "SELECT col_0 FROM table_1
    WHERE col_0 = value_0"},
    {"template": "SELECT col_0 FROM table_1
    WHERE col_1 > value_0"},
    ...
  ]
}
The "templates" list must contain exactly {num_skeletons} items.

Now, apply the same transformation to the question below. Do 
not let
specific table names, column names, or constant values
(like "description", "name", "GALAXY", or "BROADLINE")
appear in the template.

Input:
{"question": "{question.question_text}"}
Schema:
{schema_str}
Output in JSON format:
{
  "templates": [
    {"template": "..."},
    ...
  ]
}
The "templates" list must contain exactly {num_skeletons} items.
\end{lstlisting}
\end{tcolorbox}
\label{fig:dsqg_syn_skeleton_generation}

\begin{tcolorbox}[
    title=DSQG-Syn SQL Generation Prompt,
    colback=white,
    colframe=gray!75!black,
    fonttitle=\bfseries,
    enhanced,
    breakable
]
You are an expert in a specific domain and a PostgreSQL SQL expert.

\vspace{1em}

\#\#\# Instruction:

You are provided with:

1. An SQL query template.
2. A question that the query needs to answer.
3. The schema of the relevant database.
4. Optional sample values from the database columns.

You must:

1. Use only the provided schema.
2. Use only foreign-key-valid join predicates.
3. Use only provided sample values for literal filters.
4. Respect type safety for numeric and non-numeric columns.
5. Output JSON only.

\vspace{1em}

\#\#\# Prompt Body:

\begin{lstlisting}
You are an expert in a specific domain and a PostgreSQL SQL 
expert. 
You are provided with:
1. An SQL query template.
2. A question that the query needs to answer.
3. The schema of the relevant database.{sample_values_str}

Your task is to:
1. Strictly use the information from the provided schema to 
complete PostgreSQL queries. Ensure that all necessary 
table names, column names, and clauses (such as FROM and JOIN) 
come from the schema only.

2. **CRITICAL JOIN RULE**: Use ONLY foreign-key-valid join
predicates. A JOIN condition must exactly match one of the
allowed FK relationships listed below (direction can be 
reversed). Do NOT join semantically unrelated IDs 
(e.g., STATE_ID = STATION_ID) just because data types 
match.

3. **CRITICAL**: When using literal values in WHERE, HAVING, IN,
or other filter clauses, you MUST use ONLY the sample values 
provided above. Do NOT make up or hallucinate values. This 
ensures the generated queries will return actual results when 
executed against the database.

4. **CRITICAL TYPE RULE**: Use numeric operators/aggregates 
only on numeric columns.
   - AVG/SUM require numeric columns.
   - Numeric comparisons (>, >=, <, <=) require numeric/date 
   columns.
   - For text columns, use equality/inequality, IN, LIKE, 
   IS NULL, COUNT, GROUP BY.
   - Do not cast text columns to numeric unless values are 
   guaranteed numeric in schema context.
   
5. Avoid introducing any table names, column names, or other 
elements that are not explicitly defined in the schema.

6. Generate {num_sqls} PostgreSQL SQL queries that are directly 
related to the given question and fit the SQL query template.

7. Use PostgreSQL-compatible syntax only.

8. Keep the output in JSON format.

{difficulty_instructions}

Allowed FK relationships for JOINs:
{fk_constraints_str}

Numeric columns (safe for AVG/SUM and numeric comparisons):
{numeric_cols_str}

Non-numeric columns (do NOT use AVG/SUM or numeric comparisons):
{non_numeric_cols_str}

Example:
Input:
SQL Query Template:
SELECT col_1, col_2 FROM table_1 JOIN table_0 
WHERE col_3 = value_0;
Question:
What are the names and descriptions of the different types of 
photos associated with objects in the astrophysical 
classifications 
from the specobj table?
Database Schema:
CREATE TABLE photo_type (
    value number,
    name text,
    description text,
    primary key (value)
);

CREATE TABLE specobj (
    specobjid number,
    bestobjid number,
    survey text,
    class text,
    subclass text,
    primary key (specobjid),
    foreign key (bestobjid) references photoobj(objid)
);

Sample Values Available:
Table: specobj
  - class: ['GALAXY', 'STAR', 'QSO']
  - subclass: ['BROADLINE', 'STARFORMING', 'STARBURST']
  - survey: ['boss', 'sdss', 'eboss']

Output:
{
  "queries": [
    "SELECT p.name, p.description FROM photo_type p JOIN 
    specobj s ON p.value = s.bestobjid WHERE s.class = 'STAR';",
    "SELECT p.name, p.description FROM photo_type p JOIN 
    specobj s ON p.value = s.bestobjid
    WHERE s.subclass = 'BROADLINE';",
    "SELECT p.name, p.description FROM photo_type p JOIN 
    specobj s ON p.value = s.bestobjid
    WHERE s.class = 'GALAXY';"
  ]
}
Note: The WHERE clause values ('STAR', 'BROADLINE', 'GALAXY') 
are taken from the Sample Values provided.

Now, it's your turn.
Input:
SQL Query Template:
{skeleton.template}
Question:
{question.question_text}
Database Schema:
{schema_str}
Output in JSON format:
{
  "queries": [
    "..."
  ]
}
\end{lstlisting}
\end{tcolorbox}
\label{fig:dsqg_syn_sql_generation}

\begin{tcolorbox}[
    title=DSQG-Syn NLQ Synthesis Prompt,
    colback=white,
    colframe=gray!75!black,
    fonttitle=\bfseries,
    enhanced,
    breakable
]
You are an expert Data Scientist specializing in Text-to-SQL dataset curation. Your goal is to transform a SQL query into a high-fidelity Natural Language Question (NLQ).

\vspace{1em}

\#\#\# Instruction:

1. Do not leak internal SQL logic.
2. Make the NLQ sound natural.
3. Preserve the functional intent.
4. Output JSON only.

\vspace{1em}

\#\#\# Prompt Body:

\begin{lstlisting}
You are an expert Data Scientist specializing in Text-to-SQL 
dataset curation. Your goal is to transform a SQL query into a 
high-fidelity Natural Language Question (NLQ).

### NATURALNESS GUIDELINES:
1. **Selection Conciseness:** You may not list every single 
column from the `SELECT` clause if a collective term 
(e.g., "details","information","profile") is more natural.

2. **Implicit Filters:** Integrate filter criteria naturally as 
adjectives or qualifiers (e.g., "rural schools") rather than 
literal mappings (e.g., "schools where the location is 
'Rural'").

3. **Intent-based CTEs:** For queries using CTEs or complex 
subqueries, describe the *functional intent* (e.g., "For the 
most recently recorded data...") rather than the 
*execution logic* (e.g., "Find the maximum 
year and then...").

4. **Varied Phrasing:** Use a mix of questions, commands 
("List all..."), and requests ("Show the...") to maintain 
variety.

5. **No Logic Leakage:** Ensure the question does not 
explicitly "leak" the internal SQL structure (like JOIN 
conditions or specific table aliases). Use domain terminology.

### EXAMPLES:

#### Example 1 (Easy: Single Table, Simple Filter)
Input SQL: "SELECT STATION_NAME, TYPE_OF_WATER_BODY FROM 
DIM_STATION WHERE STATE_ID = 'ST_001' 
AND TYPE_OF_WATER_BODY = 'LAKE'"
Output JSON: { "question": "What are the names and water body 
types of all stations located near lakes in the first state?" }

#### Example 2 (Medium: Join, Aggregation, Group By)
Input SQL: "SELECT T1.STATE_NAME, AVG(T3.MAX_TEMPERATURE_C)
FROM DIM_STATE AS T1 JOIN DIM_STATION AS T2 ON
T1.STATE_ID = T2.STATE_ID JOIN FACT_THERMAL AS T3 ON
T2.STATION_ID = T3.STATION_ID GROUP BY T1.STATE_NAME"
Output JSON: { "question": "Show the average maximum 
temperature for each state based on available thermal
station data." }

#### Example 3 (Hard: CTE, Multiple Joins, Specific Filter)
Input SQL: "WITH top_districts AS (SELECT district_id FROM
fact_census WHERE population > 1000000) SELECT 
d.district_name, s.school_name, s.total_students FROM 
top_districts td JOIN dim_district d ON 
td.district_id = d.district_id JOIN dim_school s ON 
d.district_id = s.district_id WHERE s.school_type = 'Secondary'"
Output JSON: { "question": "For districts with a population 
over one million, list the names of secondary schools 
along with their total student counts." }

### TASK:
Input SQL: "{sql}"
Output JSON:
{
  "question": "<your natural language question>"
}
\end{lstlisting}
\end{tcolorbox}
\label{fig:dsqg_syn_nlq_synthesis}

\Needspace{8\baselineskip}
\subsection{Translation Prompts}
\label{fig:translation-prompts}
\begin{tcolorbox}[
    title=Translation System Prompt,
    colback=white,
    colframe=gray!75!black,
    fonttitle=\bfseries,
    enhanced,
    breakable
]
You are a professional translator specializing in \{target\_language\}.

\vspace{1em}

\#\#\# Instruction:

Translate the following Text-to-SQL natural language question into \{target\_language\}, following these rules:

\begin{itemize}
    \item Do NOT translate table names, column names, or any SQL-specific identifiers; keep them exactly as given in English.
    \item Translate database literal values (e.g., filter constants such as state, district, or crop names) into natural \{target\_language\} equivalents.
    \item Keep all numbers, dates, and units unchanged from the source text.
    \item Transliterate proper nouns and named entities according to standard \{target\_language\} convention; do not coin new terms.
    \item Render administrative-hierarchy terms (e.g., district, sub-district, block) using the standard census-hierarchy term in \{target\_language\}.
\end{itemize}

Return only the \{target\_language\} translation with no explanations.

\#\#\# Text:

\{question\_text\}
\end{tcolorbox}
\label{fig:translation_system}

\begin{tcolorbox}[
    title=Hinglish Translation User Prompt,
    colback=white,
    colframe=gray!75!black,
    fonttitle=\bfseries,
    enhanced,
    breakable
]
Translate this English Text-to-SQL prompt into natural Hinglish using Roman script. Keep all table names, column names, and SQL-specific values in their original English. Only translate the natural language intent and the conversational structure. Keep it technical but fluid.

\vspace{1em}

\#\#\# Text:

\{question\_text\}
\end{tcolorbox}
\label{fig:hinglish_translation_user}

\Needspace{8\baselineskip}
\subsection{Schema Generation Prompts}
\label{fig:schema-gen-prompts}
\begin{tcolorbox}[
    title=Schema Architect Prompt,
    colback=white,
    colframe=gray!75!black,
    fonttitle=\bfseries,
    enhanced,
    breakable
]
Role: Senior Database Architect.

\vspace{1em}

\#\#\# Task:

Analyze these CSV columns with their 0-based indices:

\{indexed\_columns\}

\vspace{1em}

\#\#\# Requirements:

CRITICAL PRIORITY: Domain Isolation.

Instead of one massive fact table, you must divide the data into 4--10 distinct THEMATIC TABLES based on the categories or domains of the data.\\

1. Each table should represent a single cohesive domain.\\
2. No table should have more than 12--15 columns.\\
3. Every table must have a primary key.\\
4. Link tables via foreign keys.\\
5. Ensure the total columns across all tables range between 40--80.\\

Naming conventions:

1. Use `UPPERCASE\_WITH\_UNDERSCORES` for all table and column names.\\
2. Dimension tables must start with `DIM\_`.\\
3. Fact tables must start with `FACT\_`.\\
4. Column names should be descriptive with underscores.\\

Important: Do not include indexing recommendations.
\end{tcolorbox}
\label{fig:schema_architect}

\begin{tcolorbox}[
    title=Schema Auditor Prompt,
    colback=white,
    colframe=gray!75!black,
    fonttitle=\bfseries,
    enhanced,
    breakable
]
Role: Database Normalization \& Domain Auditor.

\vspace{1em}

\#\#\# Input Schema:

\{draft\_schema\}

\vspace{1em}

\#\#\# Audit Task:

1. Width Check: Does any single table contain more than 15 columns?\\
2. Cohesion Check: Are there columns in a table that do not belong to its theme?\\
3. 3NF Violation Check: Are there transitive dependencies?\\
4. Complexity Check: Will answering benchmark questions require joining at least 3 tables?\\

List specific clumping errors and normalization failures for the architect.
\end{tcolorbox}
\label{fig:schema_auditor}

\begin{tcolorbox}[
    title=Schema Refiner Prompt,
    colback=white,
    colframe=gray!75!black,
    fonttitle=\bfseries,
    enhanced,
    breakable
]
Role: Lead Architect.

\vspace{1em}

\#\#\# Inputs:

Original Draft: \{draft\_schema\}

Auditor Feedback: \{audit\_feedback\}

Original CSV columns (with indices): \{indexed\_columns\}

\vspace{1em}

\#\#\# Task:

Resolve auditor warnings by aggressively splitting wide tables into thematic sub-domains.

1. If a table is too wide, split it logically.
2. Ensure every table has a clear join path to others.
3. Clean column names for the schema.

\vspace{1em}

\#\#\# Required Output Structure:

1. `\#\# DIMENSION TABLES`\\
2. Table definitions\\
3. `\#\# FACT TABLES`\\
4. Table definitions\\
5. `\#\# COLUMN MAPPING`

The column mapping must use exact original column names and exact source indices.
\end{tcolorbox}
\label{fig:schema_refiner}

\Needspace{8\baselineskip}
\subsection{Zero Shot Approach Prompts}
\begin{tcolorbox}[
    title=Zero-Shot System Prompt,
    colback=white,
    colframe=gray!75!black,
    fonttitle=\bfseries,
    enhanced,
    breakable
]
You are a professional database administrator and SQL expert.

\vspace{1em}

\#\#\# Instruction:

Your task is to translate a natural language question into a syntactically correct PostgreSQL query based on the provided database schema.

\vspace{1em}

\#\#\# Language and Translation Rules:

1. The input question may be in English or an Indic language.
2. You must understand the question intent and generate SQL over the English database schema.
3. If the question contains entity names in an Indic language, implicitly translate or transliterate them to match the exact English string literals found in the database schema or sample data.

\vspace{1em}

\#\#\# PostgreSQL Rules:

1. Do not use double quotes for identifiers unless strictly required.
2. Always use single quotes for string literals.
3. Cast data types explicitly if needed using `::`.
4. Output only the final SQL query.
5. Do not wrap the answer in Markdown unless explicitly requested by the task prompt.
\end{tcolorbox}
\label{fig:oneshot_system}

\begin{tcolorbox}[
    title=One-Shot User Prompt,
    colback=white,
    colframe=gray!75!black,
    fonttitle=\bfseries,
    enhanced,
    breakable
]
\#\#\# Database Schema:

\{ddl\}

\vspace{1em}

\#\#\# Sample Data:

\{samples\}

\vspace{1em}

\#\#\# One-Shot Learning Example:

\{one\_shot\}

\vspace{1em}

\#\#\# Task:

Question: \{question\}

Evidence / External Knowledge: \{evidence\}

\vspace{1em}

Output only the valid PostgreSQL query ending with a semicolon. Do not include markdown formatting.
\end{tcolorbox}
\label{fig:oneshot_user}

\end{document}